\documentclass[conference]{IEEEtran}
\usepackage{times}

\usepackage{amsmath,amsfonts,bm}

\def\eqref#1{equation~\ref{#1}}

\def\1{\bm{1}}

\DeclareMathAlphabet{\mathsfit}{\encodingdefault}{\sfdefault}{m}{sl}
\SetMathAlphabet{\mathsfit}{bold}{\encodingdefault}{\sfdefault}{bx}{n}

\DeclareMathOperator*{\argmin}{arg\,min}

\usepackage[numbers]{natbib}
\usepackage{multicol}
\usepackage[bookmarks=true]{hyperref}

\usepackage{amsfonts}
\usepackage{mathtools}
\usepackage{amsmath}
\usepackage{algorithm}
\usepackage{algpseudocode}
\usepackage{cleveref}
\usepackage{graphicx}
\usepackage{subcaption}
\usepackage{mathtools}
\usepackage{booktabs}
\usepackage{multirow}
\usepackage{xcolor}         \usepackage{xcolor-material}
\usepackage{soul}
\usepackage[makeroom]{cancel}
\let\labelindent\relax
\usepackage[shortlabels]{enumitem}
\usepackage[absolute]{textpos}

\usepackage{roboto}

\AtBeginEnvironment{appendices}{\crefalias{section}{appendix}}
\AtBeginEnvironment{appendices}{\crefalias{subsection}{appendix}}
\Crefname{figure}{Fig.}{Figs.}

\newcommand{\subjto}{\mathrm{s.t.}}
\renewcommand{\eqref}[1]{(\ref{#1})}
\newtheorem{Remark}{Remark}
\newtheorem{Theorem}{Theorem}
\newtheorem{Lemma}{Lemma}
\newtheorem{Proposition}{Proposition}

\definecolor{newColor}{HTML}{D80000}

\usepackage[framemethod=TikZ]{mdframed}
\mdfdefinestyle{ThmFrame}{linecolor=white,
    outerlinewidth=0pt,
    roundcorner=2pt,
    leftmargin=-3pt,
    rightmargin=-2pt,
    innertopmargin=5pt, innerbottommargin=5pt, innerrightmargin=5pt,
    innerleftmargin=5pt,
    backgroundcolor=MaterialBlueGrey100!40}
\mdfdefinestyle{ProofFrame}{linecolor=white,
    outerlinewidth=0pt,
    roundcorner=2pt,
    leftmargin=-3pt,
    rightmargin=-2pt,
    innertopmargin=5pt, innerbottommargin=5pt, innerrightmargin=5pt,
    innerleftmargin=5pt,
    backgroundcolor=MaterialBlueGrey100!40}
\mdfdefinestyle{AlertFrame}{linecolor=white,
    outerlinewidth=0pt,
    roundcorner=2pt,
    leftmargin=-3pt,
    rightmargin=-2pt,
    innertopmargin=5pt, innerbottommargin=5pt, innerrightmargin=5pt,
    innerleftmargin=5pt,
    backgroundcolor=MaterialRed100!40}
\mdfdefinestyle{Frame}{linecolor=white,
    outerlinewidth=0pt,
    roundcorner=2pt,
    leftmargin=-3pt,
    rightmargin=-2pt,
    innertopmargin=5pt, innerbottommargin=5pt, innerrightmargin=5pt,
    innerleftmargin=5pt,
    backgroundcolor=MaterialBlueGrey100!15}

\newcommand\extrafootertext[1]{\bgroup
    \renewcommand\thefootnote{\fnsymbol{footnote}}\renewcommand\thempfootnote{\fnsymbol{mpfootnote}}\footnotetext[0]{#1}\egroup
}

\begin{document}

\begin{textblock}{9}(1.412, 0.343)
{
\roboto
\noindent\textbf{\textsf{Robotics: \hspace{.09ex}Science and Systems 2025}} \\
\noindent\textbf{\textsf{Los Angeles, California, June 21-June 25, 2025}}
}
\end{textblock}

\title{Solving Multi-Agent Safe Optimal Control with Distributed Epigraph Form MARL}

\author{
\authorblockN{Songyuan Zhang\authorrefmark{1}\authorrefmark{2},
Oswin So\authorrefmark{1}\authorrefmark{2},
Mitchell Black\authorrefmark{3}, 
Zachary Serlin\authorrefmark{3} and
Chuchu Fan\authorrefmark{2}}
\authorblockA{\authorrefmark{2}Department of Aeronautics and Astronautics, MIT, Cambridge, Massachusetts, USA\\
Email: \texttt{\{szhang21, oswinso, chuchu\}@mit.edu}}
\authorblockA{\authorrefmark{3}MIT Lincoln Laboratory, Lexington, Massachusetts, USA\\
Email: \texttt{\{mitchell.black, Zachary.Serlin\}@ll.mit.edu}}
}

\definecolor{teaserGreen}{HTML}{00BF66}
\definecolor{teaserBlue}{HTML}{3F94CC}
\definecolor{teaserRed}{HTML}{990000}
\definecolor{teaserYellow}{HTML}{BDB76B}
\definecolor{teaserObsRed}{HTML}{FF0000}

\makeatletter
\let\@oldmaketitle\@maketitle
    \renewcommand{\@maketitle}{\@oldmaketitle
    \centering
    \vspace{0em}
    \includegraphics[trim={1cm 0 0 0},clip,width=0.99\textwidth]{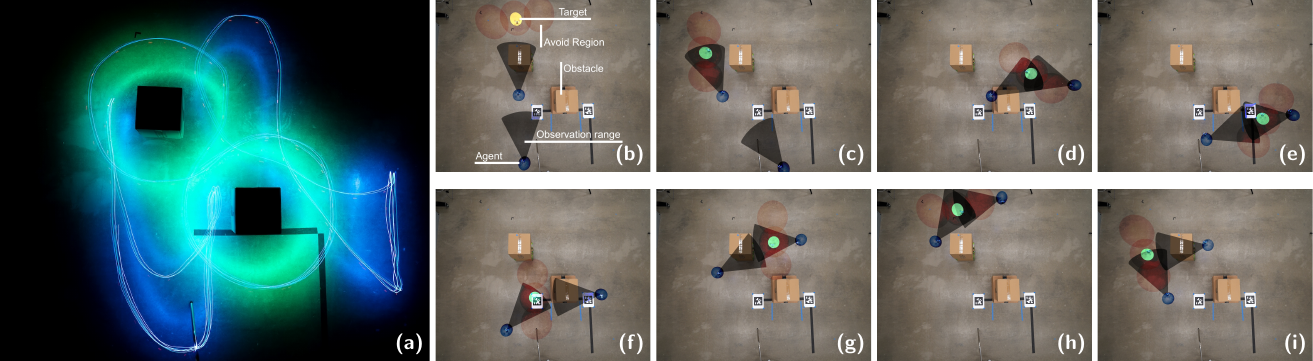}
\captionof{figure}{\textbf{Two agents using \texttt{Def-MARL} to safely and collaboratively inspect a moving target.} We propose a novel safe MARL algorithm, \texttt{Def-MARL}, that solves the multi-agent safe optimal control problem. \texttt{Def-MARL} translates the original problem to its epigraph form to avoid unstable training and extends the epigraph form to the CTDE paradigm for distributed execution.
    \textbf{(a):} Long exposure photo of the trajectories of the drones.
    The trajectory of the target is shown in \textcolor{teaserGreen}{green} and that of the agents is shown in \textcolor{teaserBlue}{blue}.
\textbf{(b)-(i):} Snapshots of the agents' policy.
    Using \texttt{Def-MARL}, the agents learn to collaborate to maintain visual contact with the target at all times, with each agent being responsible only when the target is on their side.
} 
    \vspace{-0.2cm} \label{fig: teaser}
    \setcounter{figure}{1}
}
\makeatother

\maketitle

\extrafootertext{$^*$Equal contribution. }
\begin{abstract}
Tasks for multi-robot systems often require the robots to collaborate and complete a team goal while maintaining safety.
This problem is usually formalized as a constrained Markov decision process (CMDP), which targets minimizing a global cost and bringing the mean of constraint violation below a user-defined threshold.
Inspired by real-world robotic applications, we define safety as zero constraint violation.
While many safe multi-agent reinforcement learning (MARL) algorithms have been proposed to solve CMDPs, these algorithms suffer from unstable training in this setting.
To tackle this, we use the epigraph form for constrained optimization to improve training stability and prove that the centralized epigraph form problem can be solved in a distributed fashion by each agent.
This results in a novel centralized training distributed execution MARL algorithm named \texttt{Def-MARL}.
Simulation experiments on 8 different tasks across 2 different simulators show that \texttt{Def-MARL} achieves the best overall performance, satisfies safety constraints, and maintains stable training.
Real-world hardware experiments on Crazyflie quadcopters demonstrate the ability of \texttt{Def-MARL} to safely coordinate agents to complete complex collaborative tasks compared to other methods.\footnote{Project website: \url{https://mit-realm.github.io/def-marl/}.}
\extrafootertext{\tiny{DISTRIBUTION STATEMENT A. Approved for public release. Distribution is unlimited.}}
\end{abstract}

\IEEEpeerreviewmaketitle

\section{Introduction}
Multi-agent systems (MAS) play an integral role in our aspirations for a more convenient future with examples such as autonomous warehouse operations \citep{kattepur2018distributed}, large-scale autonomous package delivery \citep{ma2017lifelong}, and traffic routing \citep{wu2020multi}.
These tasks often require designing distributed policies for agents to complete a team goal collaboratively while maintaining safety. 
To construct distributed policies, multi-agent reinforcement learning (MARL) under the centralized training distributed execution (CTDE) paradigm \citep{zhang2021multi,garg2024learning} has emerged as an attractive method. 
To incorporate the safety constraints, most MARL algorithms either choose to carefully design their objective function to incorporate \textit{soft} constraints \citep{sunehag2017value,rashid2020monotonic,yang2020qatten,wang2020qplex,peng2021facmac,rashid2020weighted}, or model the problem using the constrained Markov decision process (CMDP) \citep{altman2004constrained}, which asks for the mean constraint violation to stay below a user-defined threshold \citep{gu2023safe,liu2021cmix,ding2023provably,lu2021decentralized,geng2023reinforcement,zhao2024multi}. 
However, real-world robotic applications always require \textit{zero} constraint violation.  
While this can be addressed by setting the constraint violation threshold to zero in the CMDP, in this setting the popular Lagrangian methods experience training instabilities which result in sharp drops in performance during training, and non-convergence or convergence to poor policies \citep{so2023solving,he2023autocost,ganai2024iterative,huang2024safedreamer}.

These concerns have been identified recently, resulting in a series of works that enforce \textit{hard} constraints \citep{zanon2020safe,zhao2021model,so2023solving,ganai2024iterative,zhang2025dgppo} using techniques inspired by Hamilton-Jacobi reachability \citep{tomlin2000game,lygeros2004reachability,mitchell2005time,margellos2011hamilton,bansal2017hamilton} in deep reinforcement learning (RL) for the \textit{single-agent} case and have been shown to improve safety compared to other safe RL approaches significantly.
However, to the best of our knowledge, theories and algorithms for safe RL are still lacking for the \textit{multi-agent} scenario, especially when policies are executed in a \textit{distributed} manner.
While single-agent RL methods can be directly applied to the MARL setting by treating the MAS as a centralized single agent, the joint action space grows exponentially with the number of agents, preventing these algorithms from scaling to scenarios with a large number of agents \citep{guestrin2002coordinated,sunehag2017value,foerster2018counterfactual}.

To tackle the problem of zero constraint violation in multi-agent scenarios with \textit{distributed} policies\footnote{In this paper, the policies are distributed if each agent makes decisions using local information/sensor data and information received via message passing with other agents \citep{garg2024learning}, although this setting is sometimes called ``decentralized'' in MARL \citep{zhang2018fully}.} while achieving high collaborative performance, we propose \textbf{D}istributed \textbf{e}pigraph \textbf{f}orm \textbf{MARL} (\texttt{Def-MARL}) (\Cref{fig: teaser}). 
Instead of considering the CMDP setting, \texttt{Def-MARL} directly tackles the multi-agent safe optimal control problem (MASOCP), whose solution satisfies zero constraint violation.
To solve the MASOCP, \texttt{Def-MARL} uses the epigraph form technique \citep{boyd2004convex}, which has previously been shown to yield better policies compared to Lagrangian methods in the single-agent setting \citep{so2023solving}.
To adapt to the multi-agent setting we consider in this work, we prove that the centralized epigraph form of MASOCP can be solved in a distributed fashion by each agent. Using this result, \texttt{Def-MARL} falls under the CTDE paradigm.

We validate \texttt{Def-MARL} using 8 different tasks from 2 different simulators, multi-particle environments (MPE) \citep{lowe2017multi} and Safe Multi-agent MuJoCo \citep{gu2023safe}, with varying numbers of agents, and compare its performance with existing safe MARL algorithms using the penalty and Lagrangian methods.
The results suggest that \texttt{Def-MARL} achieves the best performance while satisfying safety: it is as safe as conservative baselines that achieve high safety but sacrifice performance, while matching the performance of unsafe baselines that sacrifice safety for high performance.
In addition, while the baseline methods require different choices of hyperparameters to perform well in different environments and suffer from unstable training because of zero constraint violation threshold, 
\texttt{Def-MARL} is stable in training using the same hyperparameters across all environments, indicative of the algorithm's robustness to environmental changes. 

We also perform real-world hardware experiments using the Crazyflie (CF) drones \cite{giernacki2017crazyflie} on two complex collaborative tasks and compare \texttt{Def-MARL} with both centralized and decentralized model predictive control (MPC) methods \cite{grne2013nonlinear}. The results indicate that \texttt{Def-MARL} finishes the tasks with $100\%$ safety rates and success rates, while the MPC methods get stuck in local minima or have unsafe behaviors.

To summarize, our \textbf{contributions} are presented below:
\begin{itemize}\item Drawing on prior work that addresses the training instability of Lagrangian methods in the zero-constraint violation setting, we extend the epigraph form method from single-agent RL to MARL, improving upon the training instability of existing MARL algorithms.
    \item We present theoretical results showing that the outer problem of the epigraph form can be decomposed and solved in a distributed manner during online execution. This allows \texttt{Def-MARL} to fall under the CTDE paradigm. 
    \item We illustrate through extensive simulations that, without any hyperparameter tuning, \texttt{Def-MARL} achieves stable training and is as safe as the most conservative baseline while simultaneously being as performant as the most aggressive baseline across all environments.
    \item We demonstrate on Crazyflie drones in hardware that \texttt{Def-MARL} can safely coordinate agents to complete complex collaborative tasks. \texttt{Def-MARL} performs the task better than centralized/decentralized MPC methods and does not get stuck in suboptimal local minima or exhibit unsafe behaviors.
\end{itemize}

\section{Related work}

\textbf{Unconstrained MARL.}
Early works that approach the problem of safety for MARL focus on navigation problems and collision avoidance \citep{chen2017decentralized,chen2017socially,everett2018motion,semnani2020multi}, where safety is achieved by a sparse collision penalty \citep{long2018towards}, or a shaped reward penalizing getting close to obstacles and neighboring agents \citep{chen2017decentralized,chen2017socially,everett2018motion,semnani2020multi}.
However, adding a penalty to the reward function changes the original objective function, so the resulting policy may not be optimal for the original constraint optimization problem. In addition, the satisfaction of collision avoidance constraints is not necessarily guaranteed by even the optimal policy \citep{massiani2022safe,everett2018motion,long2018towards}. 

\textbf{Shielding for Safe MARL.}
One popular method that provides safety to learning-based methods is using \textit{shielding} or a \textit{safety filter} \citep{garg2024learning}. Here, an unconstrained learning method is paired with a shield or safety filter using techniques such as predictive safety filters \citep{zhang2019mamps,muntwiler2020distributed}, control barrier functions \citep{cai2021safe,pereira2022decentralized}, or automata \citep{elsayed2021safe,xiao2023model,shield_MARL_amato,bharadwaj2019synthesis}. Such shields are often constructed before the learning begins and are used to modify either the feasible actions or the output of the learned policy to maintain safety.
One benefit is that safety can be guaranteed during both training and deployment since the shield is constructed before training. However, they require domain expertise to build a valid shield, which can be challenging in the single-agent setting and even more difficult for MAS \citep{garg2024learning}. Other methods can automatically synthesize shields but face scalability challenges \citep{shield_MARL_amato,shield_MARL_topcu}. Another drawback is that the policy after shielding might not consider the same objective as the original policy and may result in noncollaborative behaviors or deadlocks \citep{qin2021learning,zhang2023neural,zhang2024gcbf+}. 

\textbf{Constrained MARL. }
In contrast to unconstrained MARL methods, which change the constraint optimization problem to an unconstrained problem, constrained MARL methods explicitly solve the CMDP problem. For the single-agent case, prominent methods for solving CMDPs include primal methods \citep{xu2021crpo}, primal-dual methods using Lagrange multipliers \citep{borkar2005actor,tessler2018reward,he2023autocost,huang2024safedreamer}, and trust-region-based approaches \citep{achiam2017constrained,he2023autocost}. These methods provide guarantees either in the form of asymptotic convergence guarantees to the optimal (safe) solution \citep{borkar2005actor,tessler2018reward} using stochastic approximation theory \citep{robbins1951stochastic,borkar2009stochastic}, or recursive feasibility of intermediate policies \citep{achiam2017constrained,satija2020constrained} using ideas from trust region optimization \citep{schulman2015trust}.
The survey \citep{gu2022review} provides an overview of the different methods of solving safety-constrained single-agent RL. In multi-agent cases, however, the problem becomes more difficult because of the non-stationary behavior of other agents, and similar approaches have been presented only recently \citep{gu2023safe,liu2021cmix,ding2023provably,lu2021decentralized,geng2023reinforcement,zhao2024multi,chen2024on}. However, the CMDP setting they handle makes it difficult for them to handle hard constraints, and results in poor performance with zero constraint violation threshold \citep{ganai2024iterative}.

\textbf{Model predictive control. }
Distributed MPC methods have been proposed to handle MAS, incorporating multi-agent path planning, machine learning, and distributed optimization \citep{wang2014synthesis,toumieh2022decentralized,zhu2020trajectory,fedele2023distributed,luis2019trajectory,conte2012computational,nedic2018distributed}.
However, the solution quality of nonlinear optimizers used to solve MPC when the objective function and constraints are nonlinear highly depends on the initial guess \cite{tsiotras2011initial,grne2013nonlinear}.
Moreover, the real-time nonlinear optimizers typically require access to (accurate) first and second-order derivatives \cite{nocedal1999numerical,grne2013nonlinear}, which present challenges when trying to solve tasks that have non-differentiable or discontinuous cost functions and constraints such as the ones we consider in this work.

\section{Problem setting and preliminaries}

\subsection{Multi-agent safe optimal control problem}

We consider the multi-agent safe optimal control problem (MASOCP) as defined below. Consider a homogeneous MAS with $N$ agents. At time step $k$, the global state and control input are given by $x^k\in\mathcal X\subseteq\mathbb R^n$ and $u^k\in\mathcal U\subseteq\mathbb R^m$, respectively. The global control vector is defined by concatenation $u^k \coloneqq [u_1^k;\dots;u_N^k]$, where $u_i^k\in\mathcal U_i$ is the control input of agent $i$.
We consider the general nonlinear discrete-time dynamics for the MAS:
\begin{equation} \label{eq:dynamics}
    x^{k+1} = f(x^k, u^k),
\end{equation}
where $f:\mathcal X\times\mathcal U\to\mathcal X$ is the global dynamics function. 
We consider the \textit{partially observable} setting, where each agent has a limited communication radius $R>0$ and can only communicate with other agents or observe the environment within its communication region. Denote $o_i^k = O_i(x^k)\in\mathcal O\subseteq \mathbb R^{n_o}$ as the vector of the information observed by agent $i$ at the time step $k$, where $O_i:\mathcal X\to\mathcal O$ is an encoding function of the information shared from neighbors of agent $i$ and the observed data of the environment. We allow multi-hop communication between agents, so an agent may communicate with another agent outside its communication region if a communication path exists between them. 

Let the avoid/unsafe set of agent $i$ be $\mathcal A_i\coloneqq\{o_i\in\mathcal O:h_i(o_i)>0\}$, for some function $h_i:\mathcal O\to \mathbb R$. The global avoid set is then defined as $\mathcal A\coloneqq\{x\in\mathcal X:h(x)>0\}$, where $h(x) = \max_i h_i(o_i) = \max_i h_i(O_i(x))$. In other words, $\exists i, \;\mathrm{s.t.}\; o_i \in \mathcal{A}_i \iff x \in \mathcal{A}$. Given a global cost function $l: \mathcal X\times\mathcal U\to\mathbb R$ describing the task for the agents to accomplish\footnote{The cost function $l$ is \textbf{not} the cost in CMDP. Rather, it corresponds to the negation of the \textit{reward} in CMDP.}, we aim to find distributed control policies $\pi_i:\mathcal O\to\mathcal U_i$ such that starting from any given initial states $x^0\notin\mathcal A$, the policies keep the agents outside the avoid set $\mathcal{A}$ and minimize the infinite horizon cost. In other words, denoting $\pi:\mathcal X\to\mathcal U$ as the joint policy such that $\pi(x) = [\pi_1(o_1);\dots;\pi_N(o_N)] = [\pi_1(O_1(x));\dots;\pi_N(O_N(x))]$, we aim to solve the following infinite-horizon MASOCP for a given initial state $x^0$:
\begin{subequations}\label{eq: macocp}
\begin{align}
    \hspace{-1em}\min_{\{\pi_i\}_{i=1}^N} \quad& \sum_{k=0}^\infty l(x^k, \pi(x^k)) \label{eq: macocp:obj}\\
    \subjto \quad
    & h_i(O_i(x^k)) \leq 0, && \hspace{-0.4em}\forall i\in\{1,\dots,N\}, k\geq 0,\label{eq: hbar} \\
    & x^{k+1} = f(x^k,\pi(x^k)),  && k\geq 0. \label{eq: macocp:dyn}
\end{align}
\end{subequations}
Note that the safety constraint \eqref{eq: hbar} differs from the average constraints considered in CMDPs \citep{altman2004constrained}. Consequently, instead of allowing safety violations to occur as long as the mean constraint violation is below a threshold, this formulation disallows \textit{any} constraint violation. From hereon, we omit the dynamics constraint \eqref{eq: macocp:dyn} for conciseness.

\subsection{Epigraph form}\label{sec: EF}
Existing methods are unable to solve \eqref{eq: macocp} well.
This has been observed previously in the single-agent setting \citep{zanon2020safe,zhao2021model,so2023solving,ganai2024iterative}.
We show later that the poor performance of methods that tackle the CMDP setting to the constrained problem \eqref{eq: macocp} also translates to the multi-agent setting, as we observe a similar phenomenon in our experiments (\Cref{sec: experiments}).
Namely, although unconstrained MARL can be used to solve \eqref{eq: macocp} using the penalty method \citep{nayak2023scalable}, this does not perform well in practice, where a small penalty results in policies that violate constraints, and a large penalty results in higher total costs.
The Lagrangian method \citep{gu2023safe} can solve the problem theoretically, but it suffers from unstable training and has poor performance in practice when the constraint violation threshold is zero \citep{so2023solving,ganai2024iterative}.
In this section, we introduce a new method of solving \eqref{eq: macocp} that can mitigate the above problems by extending prior work \citep{so2023solving} to the multi-agent setting.

Given a constrained optimization problem with objective function $J$ (e.g., $J = \sum_{k=0}^\infty l(x^k,\pi(x^k))$ as in \eqref{eq: macocp:obj}), and constraints $h$ (e.g., \eqref{eq: hbar}):
\begin{align} \label{eq:constr_problem}
    \min_{{\pi}} \quad J({\pi}) \qquad \subjto \quad h({\pi})\leq 0,
\end{align}
its epigraph form \citep{boyd2004convex} is given as
\begin{align} \label{eq:epigraph_orig}
    \min_{{\pi},z}\quad z \qquad \subjto \quad h({\pi})\leq 0, \quad J({\pi})\leq z,
\end{align}
where $z\in\mathbb R$ is an auxiliary variable. In other words, we add a constraint to enforce $z$ as an upper bound of the cost $J(\pi)$, then minimize $z$. The solution to \eqref{eq:epigraph_orig} is identical to the original problem \eqref{eq:constr_problem} \citep{boyd2004convex}.
Furthermore, \eqref{eq:epigraph_orig} is equivalent \citep{so2023solving} to
\begin{subequations}\label{eq: ef}
\begin{align}
    \min_z &\quad z \label{eq: ef-outer}\\
    \subjto &\quad \min_{{\pi}} J_z({\pi}, z) \coloneqq \max \{h({\pi}), J({\pi}) - z\}\leq 0\label{eq: ef-inner}
\end{align}
\end{subequations}
As a result, the original constrained problem \eqref{eq:constr_problem} is decomposed into the following two subproblems:
\begin{enumerate}\item An unconstrained \textit{inner problem} \eqref{eq: ef-inner}, where, given an arbitrary desired cost upper bound $z$, we find ${\pi}$ such that $J_z({\pi}, z)$ is minimized, i.e., best satisfies the constraints $h \leq 0$ and $J \leq z$.
    \item A $1$-dimensional constrained \textit{outer problem} \eqref{eq: ef-outer} over $z$, which finds the smallest cost upper bound $z$ such that $z$ is indeed a cost upper bound ($J\leq z)$ and the constraints of the original problem $h({\pi}) \leq 0$ holds.
\end{enumerate}

\textbf{Comparison with the Lagrangian method. }
Another popular way to solve MASOCP \eqref{eq: macocp} is the Lagrangian method \citep{gu2023safe}.
However, it suffers from unstable training when considering the \textit{zero} constraint violation \citep{so2023solving,he2023autocost} setting.
More specifically, this refers to the case with constraints $\sum_{k=0}^\infty c(x^k) \leq 0$ for $c : \mathcal{X} \to \mathbb{R}_{\geq 0}$ non-negative.
Since $h$ can be negative, we can convert our problem setting \eqref{eq:constr_problem} to the zero constraint violation setting by taking $c(x) \coloneqq \max\{ 0, h(x) \}$.
Then, \eqref{eq:constr_problem} reads as
\begin{equation} \label{eq: constr_new}
    \min_{\pi} \quad J(\pi) \qquad \subjto \quad \sum_{k=0}^\infty \max\{ 0, h(x^k) \} \leq 0.
\end{equation}
The Lagrangian form of \eqref{eq: constr_new} is then
\begin{equation}
    \max_{\lambda\geq0} \min_{\pi} \quad J_\lambda(\pi, \lambda) \coloneqq J(\pi) + \lambda \sum_{k=0}^\infty \max\{h(x^k), 0\},
\end{equation}
where $\lambda$ is the Lagrangian multiplier and is updated with gradient ascent. 
However, $\frac{\partial}{\partial\lambda}J_\lambda({\pi}, \lambda) = \sum_{k=0}^\infty \max\{h(x^k), 0\}\geq 0$, so $\lambda$ continuously increases and never decreases.
As $\frac{\partial}{\partial {\pi}}J_\lambda({\pi}, \lambda)$ scales linearly in $\lambda$ when $h(x^k)>0$ for some $k$, a large value of $\lambda$ causes a large gradient w.r.t $x$, and makes the training unstable. Note that for the epigraph form, since $z$ does not \textit{multiply} with the cost function $J$ but is \textit{added} to $J$ in \eqref{eq: ef-inner}, the gradient $\frac{\partial}{\partial {\pi}}J_z({\pi}, z)$ does not scale with the value of $z$ resulting in more stable training. We validate this in our experiments (\Cref{sec: experiments}). 

\section{Distributed epigraph form multi-agent reinforcement learning}

In this section, we propose the \textbf{D}istributed \textbf{e}pigraph \textbf{f}orm \textbf{MARL} (\texttt{Def-MARL}) algorithm to solve MASOCP \eqref{eq: macocp} using MARL. First, we transfer MASOCP \eqref{eq: macocp} to its epigraph form with an auxiliary variable $z$ to model the desired cost upper bound. The epigraph form includes an inner problem and an outer problem. 
For distributed execution, we provide a theoretical result that the outer problem can be solved distributively by each agent. This allows \texttt{Def-MARL} to fit the CTDE paradigm, where in centralized training, the agents' policies are trained together given the desired cost upper bound $z$, and in distributed execution, the agents distributively find the smallest cost upper bound $z$ that ensures safety. 

\subsection{Epigraph form for MASOCP}

To rewrite MASOCP \eqref{eq: macocp} into its epigraph form \eqref{eq: ef}, we first define the cost-value function $V^l$ for a joint policy $\pi$ using the standard optimal control notation \citep{bertsekas2012dynamic}:
\begin{equation}
    V^l(x^\tau;\pi) \coloneqq \sum_{k\geq\tau} l(x^k, \pi(x^k) ).
\end{equation}
We also define the constraint-value function $V^h$ as the maximum constraint violation:
\begin{equation}\label{eq: Vh-def}
    \begin{aligned}
        V^h(x^\tau;\pi) &\coloneqq \max_{k\geq \tau} h(x^k) = \max_{k\geq\tau}\max_{i}h_i(o_i^k)\\
        &= \max_i\max_{k\geq\tau} h_i(o_i^k) = \max_i V_i^h(o_i^\tau;\pi).
\end{aligned}
\end{equation}
Here, we interchange the $\max$ to define the \textit{local per-agent} functions $V_i^h(o_i^\tau;\pi) = \max_{k\geq\tau}h_i(o_i^k)$.
Each $V_i^h$ uses only the agent's local observation and thus is distributed.
We now introduce the auxiliary variable $z$ for the desired upper bound of $V^l$,
allowing us to restate \eqref{eq: macocp} concisely as
\begin{align}\label{eq: macocp_concise}
\min_{\{\pi_i\}_{i=1}^N} \quad V^l(x^0; \pi)
    \qquad \subjto \quad
    V^h(x^0; \pi) \leq 0.
\end{align}
The epigraph form \eqref{eq: ef} of \eqref{eq: macocp_concise} then takes the form
\begin{subequations}\label{eq: V_ef}
    \begin{align}
        \min_z &\quad z \label{eq: ef-macocp-outer-0}
        \vphantom{\underbrace{\max\big\{\big\}}_{\coloneqq V}}\\
        \subjto &\quad
        \min_{\{\pi_i\}_{i=1}^N}\, \underbrace{\max\big\{ \max_i V_i^h(o_i^\tau;\pi), V^l(x^\tau;\pi) - z \big\}}_{\coloneqq V(x^0, z; \pi)} \leq 0. \label{eq: V_ef:polmin}
    \end{align}
\end{subequations}
By interpreting the left-hand side of \eqref{eq: V_ef:polmin} as a \textit{new} policy optimization problem, we define the \textit{total} value function $V$ as the objective function to \eqref{eq: V_ef:polmin}.
This can be simplified as
\begin{equation}\label{eq: V-def}
    \begin{aligned}
    V(x^\tau,z;\pi) &= \max\{\max_i V_i^h(o_i^\tau;\pi), V^l(x^\tau;\pi) - z\} \\
    &= \max_i\max\{V_i^h(o_i^\tau;\pi), V^l(x^\tau;\pi) - z\} \\ &= \max_i V_i(x^\tau,z;\pi),
\end{aligned}
\end{equation}
Again, we interchange the $\max$ to define $V_i(x^\tau, z;\pi) = \max\{V_i^h(o_i^\tau;\pi), V^l(x^\tau;\pi) - z\}$ as the \textit{per-agent} total value function.
Using this to rewrite \eqref{eq: V_ef} then yields
\begin{subequations}\label{eq: ef-macocp}
    \begin{align}
        \min_z& \quad z \label{eq: ef-macocp-outer}
        \vphantom{\Big(}\\
        \subjto& \quad {\min_{\pi}} \max_i V_i(x^0, z;\pi)\leq 0.\label{eq: ef-macocp-inner}
        \vphantom{\Big(}
    \end{align}
\end{subequations}
This decomposes the original problem \eqref{eq: macocp}
into an unconstrained inner problem \eqref{eq: ef-macocp-inner} over policy $\pi$ and a constrained outer problem \eqref{eq: ef-macocp-outer} over $z$.
During offline training, we solve the inner problem \eqref{eq: ef-macocp-inner}: for parameter $z$, find the optimal policy $\pi(\cdot,z)$ to minimize $V(x^0,z;\pi)$.
Note that the optimal policy of the inner problem depends on $z$.
During execution, we solve the outer problem \eqref{eq: ef-macocp-outer} online to get the minimal $z$ that satisfies constraint \eqref{eq: ef-macocp-inner}. Using this $z$ in the $z$-conditioned policy $\pi(\cdot,z)$ found in the inner problem gives us the optimal policy for the overall epigraph form MASOCP (EF-MASOCP).

\begin{figure*}[t]
    \centering
    \includegraphics[width=.99\textwidth]{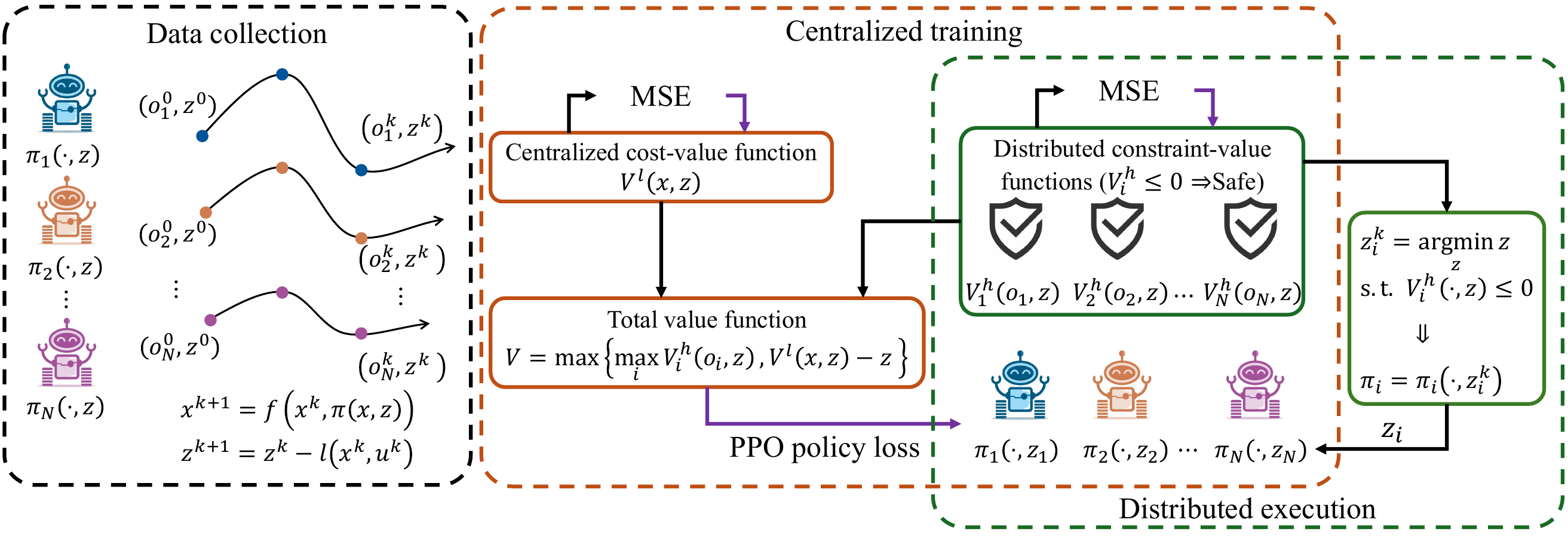}
    \caption{\textbf{\texttt{Def-MARL} algorithm. } Randomly sampled initial states and $z^0$ are used to collect trajectories in $x$ and $z$ using the current policy $\pi$.
    In the centralized training (orange blocks), distributed constraint-value functions $V^h_i$ and policies $\pi_i$ and a centralized cost-value function $V^l$ are jointly trained.
    During distributed execution (green blocks), the distributed $V^h_i$ are used to solve the outer problem \eqref{eq: dec-ef-macocp-2} to compute the optimal $z_i$, which is used in each agent's $z$-conditioned policy.
   }
    \label{fig: algorithm}
\end{figure*}

To solve the inner problem \eqref{eq: ef-macocp-inner}, the total value function $V$ must be amenable to dynamic programming, which we show in the following proposition.
\begin{Proposition}\label{thm: dynamic-program}
    Dynamic programming can be applied to EF-MASOCP \eqref{eq: ef-macocp}, resulting in
    \begin{equation}\label{eq: dynamic-program}
        \begin{aligned}
            V(x^k, z^k;\pi) &= \max\{h(x^k), V(x^{k+1},z^{k+1};\pi)\}, \\
        \quad z^{k+1} &= z^{k} - l(x^k,\pi(x^k)).
        \end{aligned}
\end{equation}
\end{Proposition}
The proof of \Cref{thm: dynamic-program} is provided in \Cref{app: proof-dp} following the proof of the single-agent version \citep{so2023solving}. In other words, for a given cost upper bound $z^k$, the value function $V$ at the current state $x^k$ can be computed using the value function at the next state $x^{k+1}$ but with a \textit{different} cost upper bound $z^{k+1} = z^k - l(x^k,\pi(x^k))$ which itself is a function of $z^k$. This can be interpreted as a ``\textit{dynamics}'' for the cost upper bound $z$.
Intuitively, if we wish to satisfy the upper bound $z^k$ but suffer a cost $l(x^k,\pi(x^k))$, then the upper bound at the next time step should be smaller by $l(x^k,\pi(x^k))$ so that the total cost from $x^k$ remains upper bounded by $z^k$.
Additional discussion on \Cref{thm: dynamic-program} is provided in \Cref{app: discussion-dp}.

\begin{Remark}[Effect of $z$ on the learned policy]
    From \eqref{eq: V-def}, for a \textit{fixed} $x$ and $\pi$, observe that for $z$ large enough (i.e., ${V^l(x; \pi)} - z$ is small enough), we have {$V(x, z; \pi) = V^h(x;\pi)$}.
Consequently, taking a gradient step on $V(x, z; \pi)$ equals taking a gradient step on $V^h(x; \pi)$, which reduces the constraint violation.
Otherwise, {$V(x, z; \pi) = V^l(x; \pi) - z$}.
{Taking gradient steps on $V(x, z; \pi)$ equals taking gradient steps on $V^l(x; \pi)$, which reduces the total cost.}
\end{Remark}

\subsection{Solving the inner problem using MARL}

Following \citet{so2023solving}, we solve the inner problem using centralized training with proximal policy optimization (PPO) \citep{schulman2017proximal}.
We use a graph neural network (GNN) backbone for the $z$-conditioned policy $\pi_\theta(o_i, z)$, cost-value function $V^l_\phi(x, z)$, and the constraint-value function $V^h_\psi(o_i, z)$ with parameters $\theta$, $\phi$, and $\psi$, respectively.
Note that other neural network (NN) structures can be used as well.
The implementation details are introduced in \Cref{app: experiments}. 

\textbf{Policy and value function updates. }
During centralized training, the NNs are trained to solve the inner problem \eqref{eq: ef-macocp-inner}, i.e., for a randomly sampled $z$, find policy $\pi(\cdot,z)$ that minimizes the total value function $V(x^0,z;\pi)$.
We follow MAPPO \citep{yu2022surprising} to train the NNs.
Specifically, when calculating the advantage with the generated advantage estimation (GAE) \citep{schulman2015high} for the $i$-th agent, $A_i$ \citep{schulman2017proximal}, instead of using the cost function $V^l$ \citep{yu2022surprising}, we apply the decomposed total value function $\max\{V^h_\psi(o_i,z), V^l_\phi(x,z) - z\}$.
We perform trajectory rollouts following the dynamics for $x$ \eqref{eq:dynamics} and $z$ \eqref{eq: dynamic-program} using the learned policy $\pi_\theta$,
starting from random sampled $x^0$ and $z^0$.
After collecting the trajectories, we train the cost-value function $V^l_\phi$ and the constraint-value function $V^h_\psi$ via regression and use the PPO policy loss to update the $z$-conditioned policy $\pi_\theta$.

\begin{figure*}[t]
    \centering
    \includegraphics[width=.99\textwidth,trim={0 0.5cm 0 0.5cm},clip]{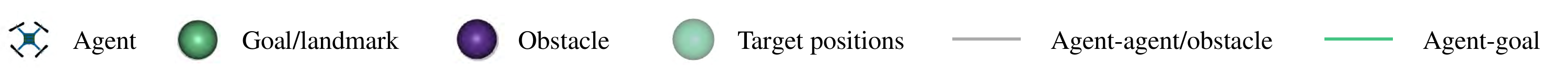}
    \captionsetup[subfigure]{labelformat=empty}
    \begin{subfigure}{.16\textwidth}
        \centering
        \includegraphics[width=\columnwidth]{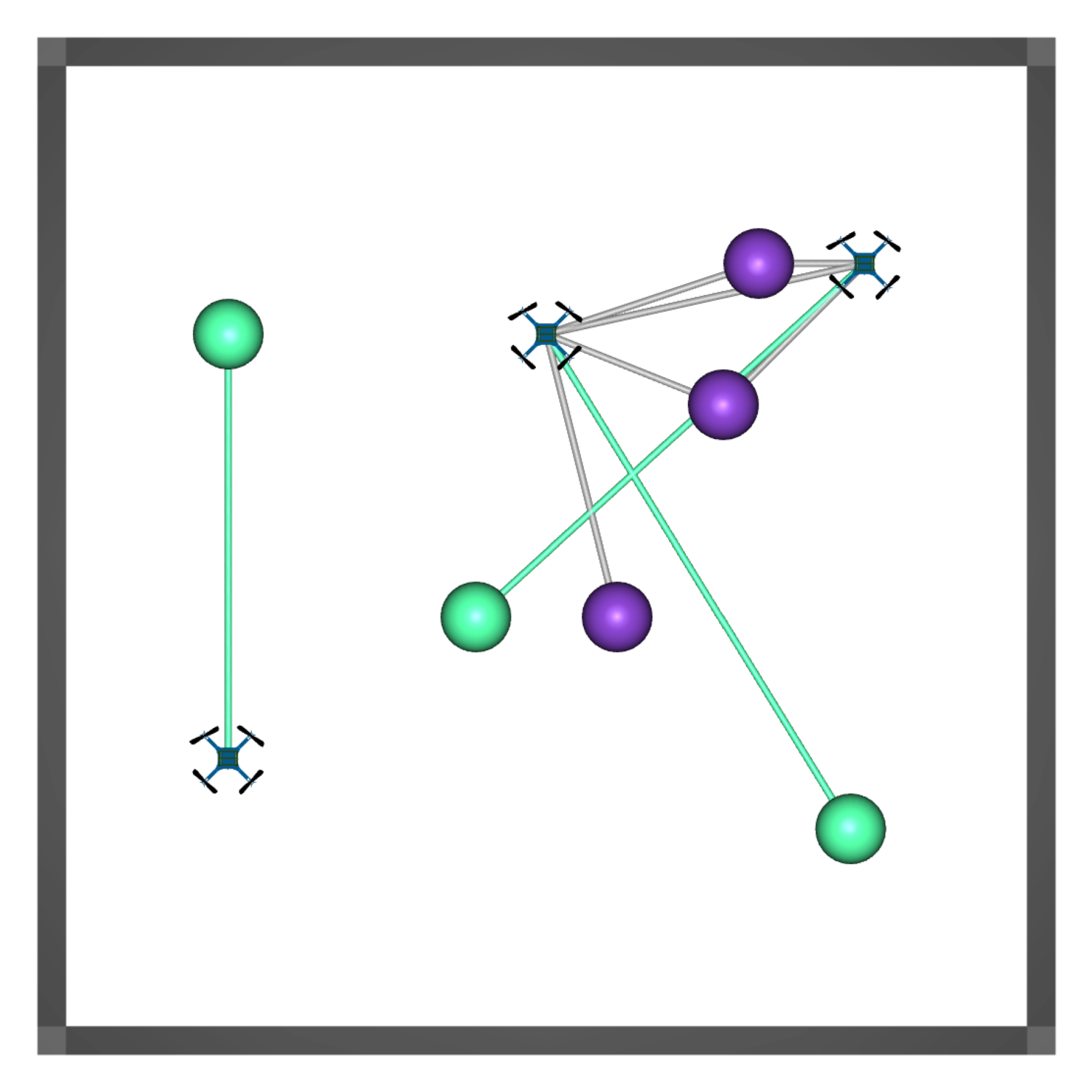}
        \caption{\textsc{Target}}
    \end{subfigure}
    \captionsetup[subfigure]{labelformat=empty}
    \begin{subfigure}{.16\textwidth}
        \centering
        \includegraphics[width=\columnwidth]{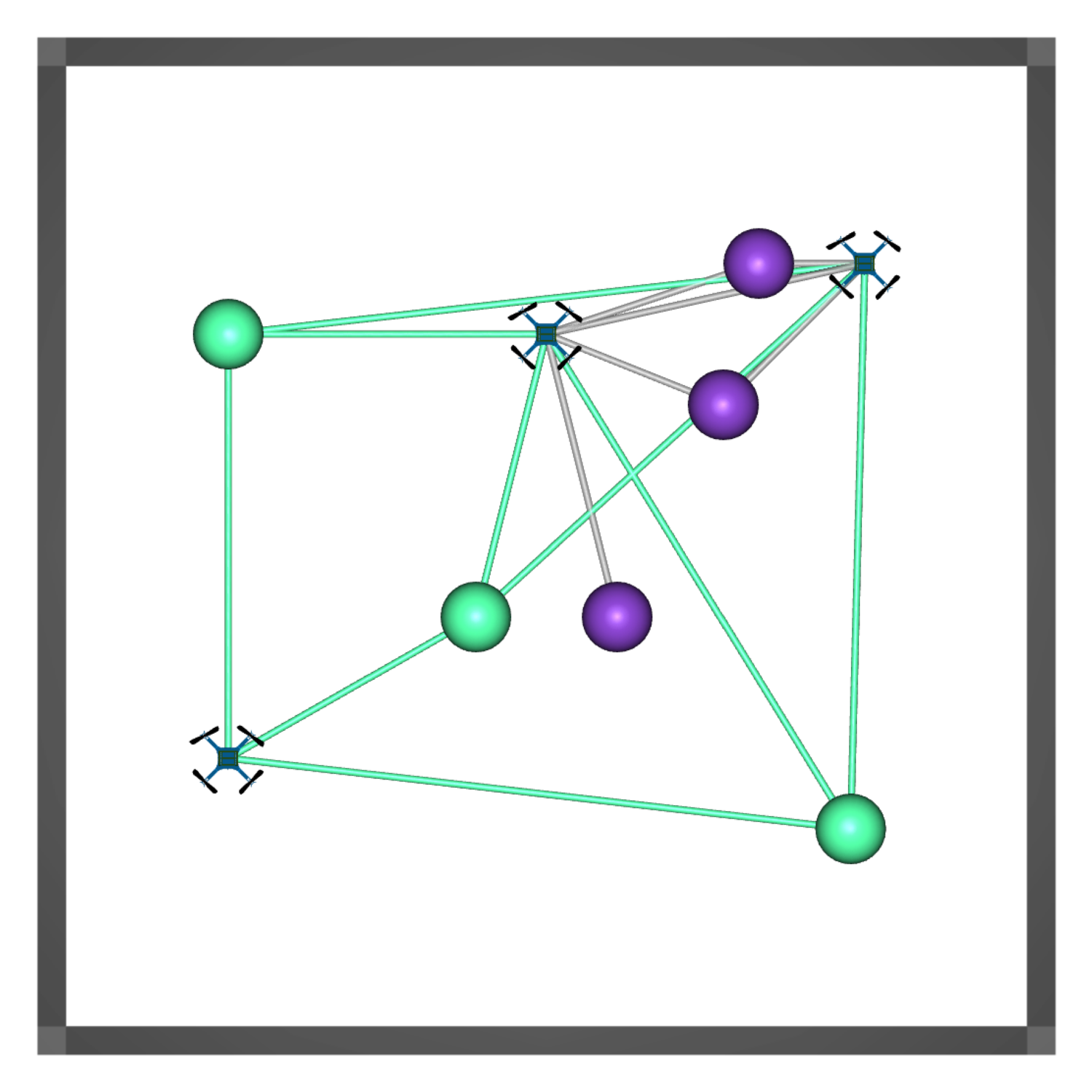}
        \caption{\textsc{Spread}}
    \end{subfigure}
    \captionsetup[subfigure]{labelformat=empty}
    \begin{subfigure}{.16\textwidth}
        \centering
        \includegraphics[width=\columnwidth]{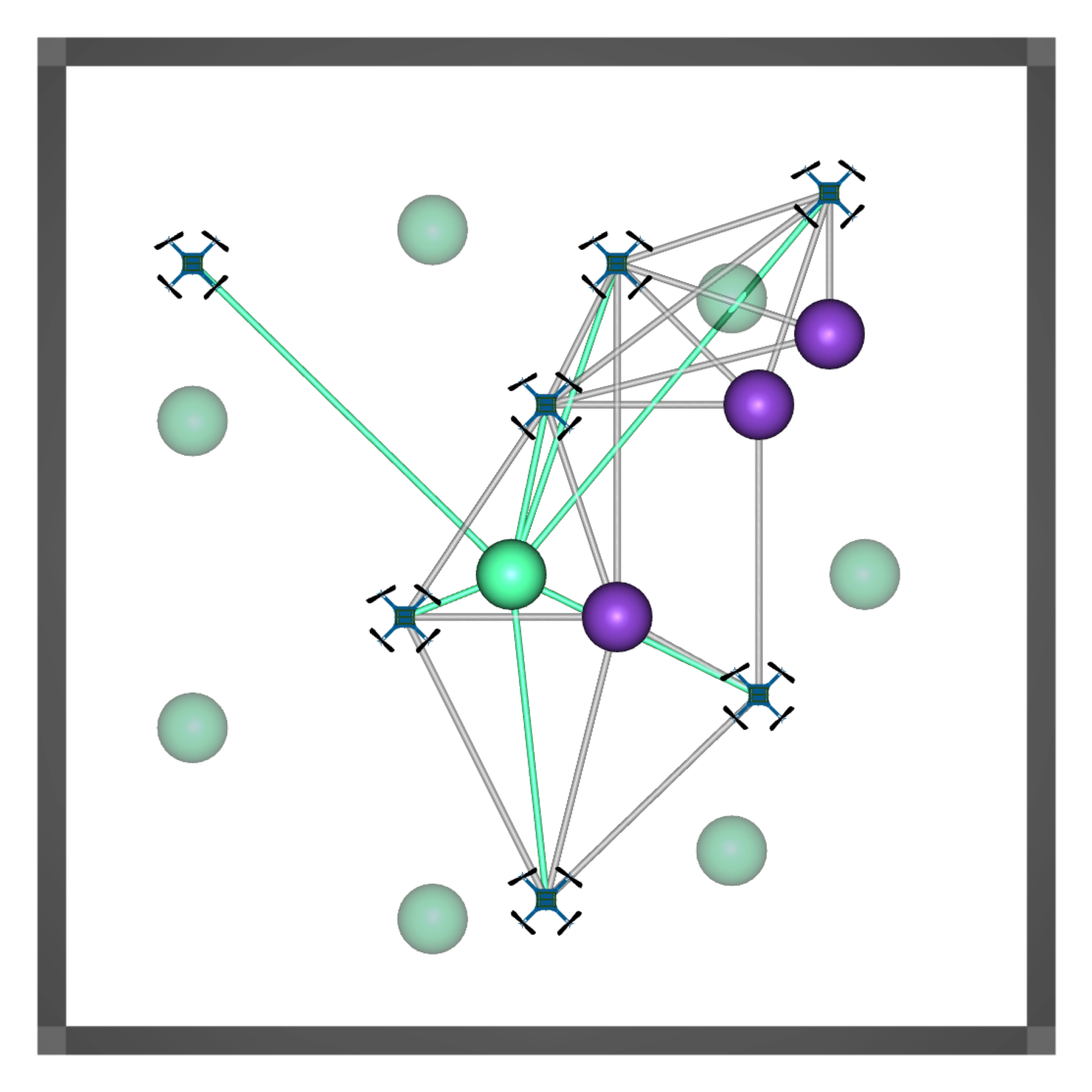}
        \caption{\textsc{Formation}}
    \end{subfigure}
    \captionsetup[subfigure]{labelformat=empty}
    \begin{subfigure}{.16\textwidth}
        \centering
        \includegraphics[width=\columnwidth]{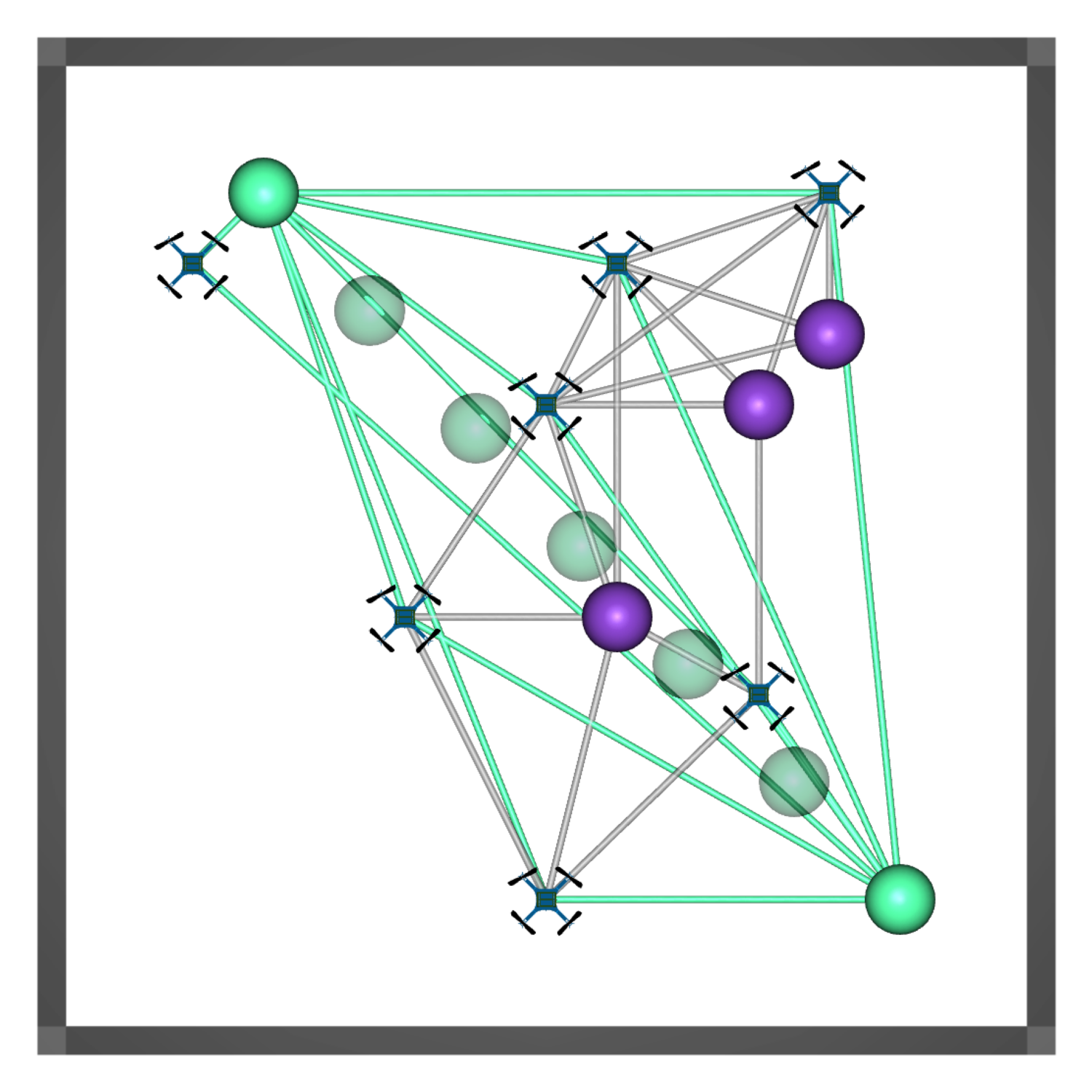}
        \caption{\textsc{Line}}
    \end{subfigure}
    \captionsetup[subfigure]{labelformat=empty}
    \begin{subfigure}{.16\textwidth}
        \centering
        \includegraphics[width=\columnwidth]{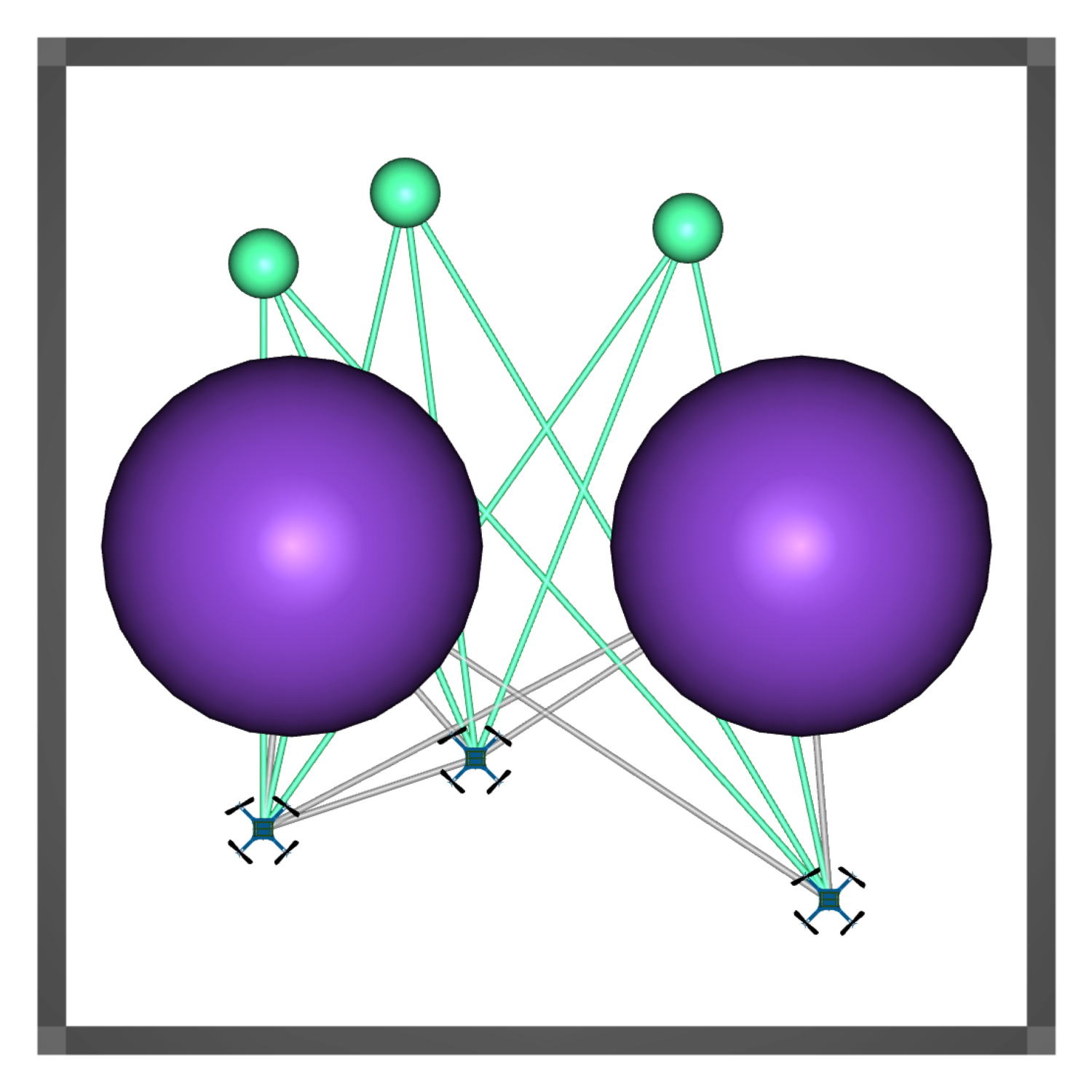}
        \caption{\textsc{Corridor}}
    \end{subfigure}
    \captionsetup[subfigure]{labelformat=empty}
    \begin{subfigure}{.16\textwidth}
        \centering
        \includegraphics[width=\columnwidth]{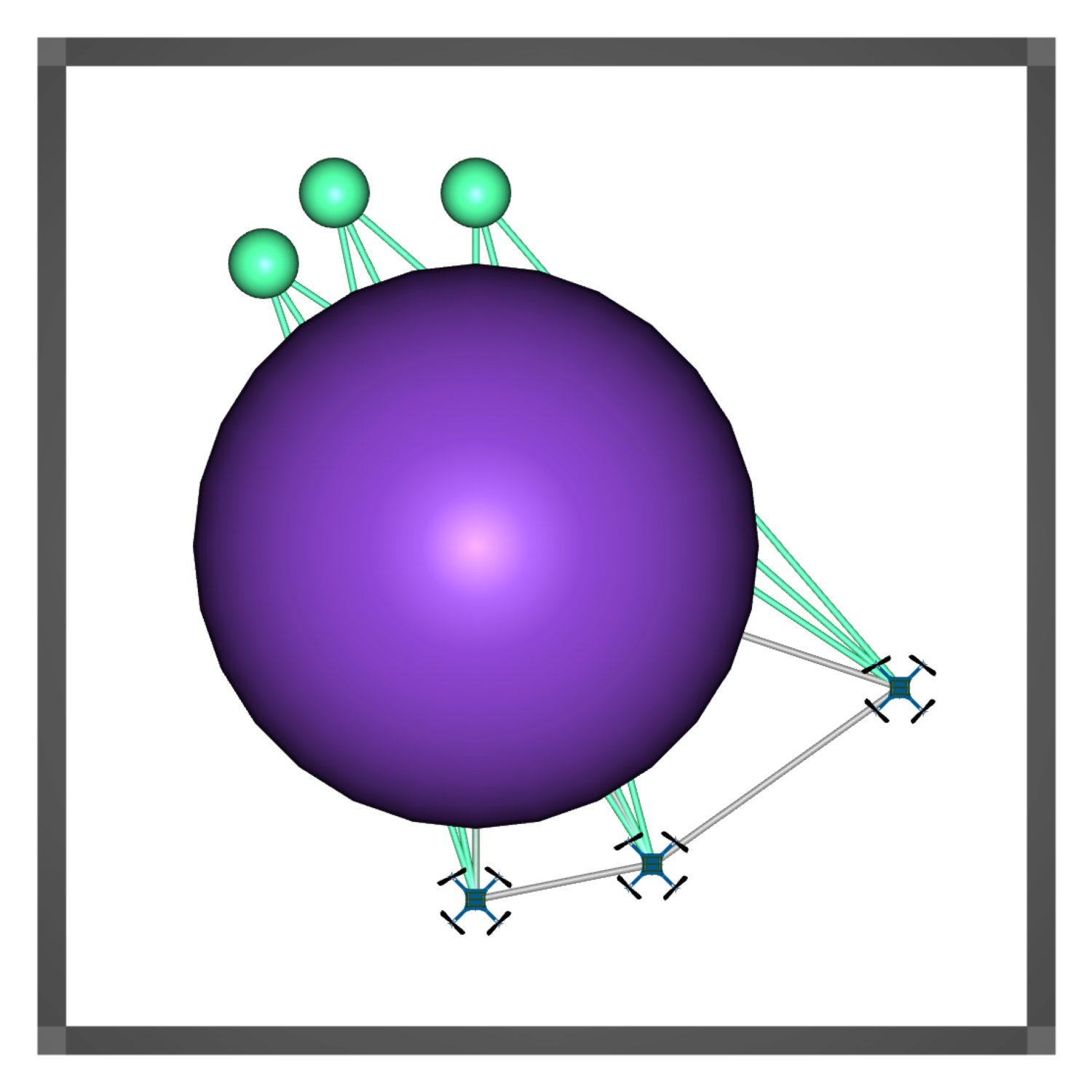}
        \caption{\textsc{ConnectSpread}}
    \end{subfigure}

    \vspace{1ex}

    \hspace{.1\linewidth}\begin{subfigure}[t]{0.4\linewidth}
        \centering
        \includegraphics[width=0.8\linewidth,trim={20cm 6cm 20cm 5cm},clip]{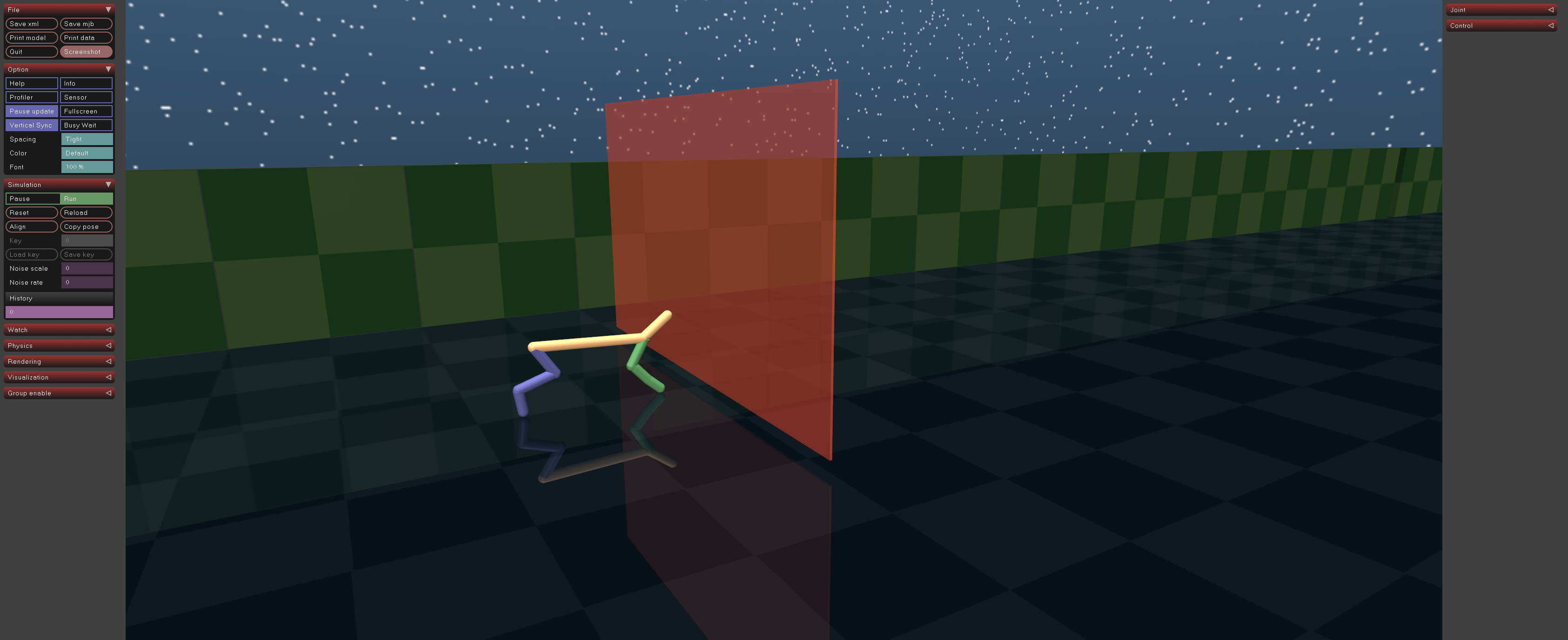}
        \caption{\textsc{Safe HalfCheetah(2x3)}}
    \end{subfigure}\begin{subfigure}[t]{0.4\linewidth}
        \centering
        \includegraphics[width=0.8\linewidth,trim={20cm 6cm 20cm 5cm},clip]{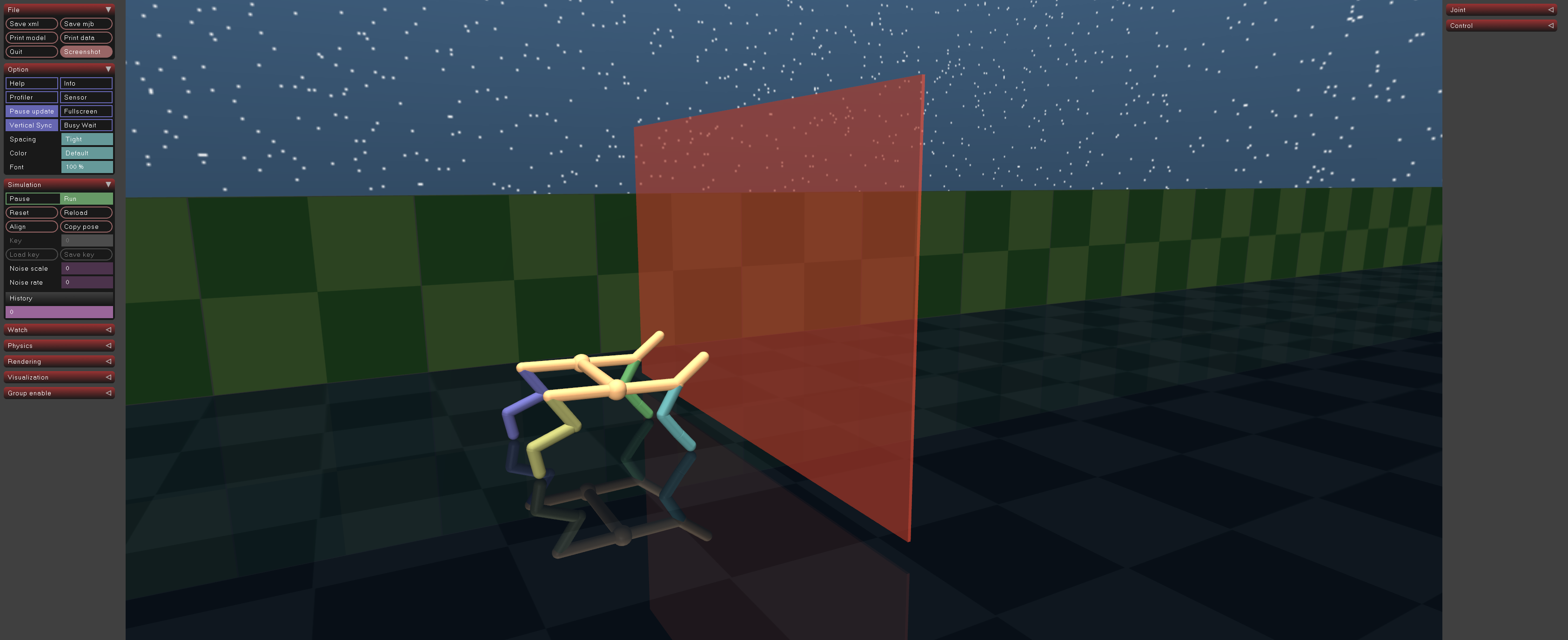}
        \caption{\textsc{Safe Coupled HalfCheetah(4x3)}}
    \end{subfigure}\hspace{.1\linewidth}

    \caption{\textbf{Simulation Environments. }
    Visualization of the (top) \textit{modified} MPE \cite{lowe2017multi} and (bottom) Safe Multi-agent MuJoCo \cite{gu2023safe} environments we consider.
}
    \label{fig: env}
\end{figure*}

\subsection{Solving the outer problem during distributed execution}\label{sec: decentralized-exe}

During execution, we solve the outer problem of EF-MASOCP \eqref{eq: ef-macocp} online. However, the outer problem is still centralized because the constraint \eqref{eq: ef-macocp-inner} requires the centralized cost-value function $V^l$. To achieve a distributed policy during execution, we introduce the following theoretical result:
\begin{Theorem}\label{thm: dec-ef-macocp}
Assume no two unique values of $z$ achieves the same unique cost.
Then, the outer problem of EF-MASOCP \eqref{eq: ef-outer} is equivalent to the following:
    \begin{subequations}\label{eq: dec-ef-macocp}
        \begin{align}
            z &= \max_i z_i\label{eq: dec-ef-macocp-1}\\
            \begin{split}
                z_i &= \min_{z'} \quad z' \\
                &\hspace{1.6em}\subjto\quad V_i^h(o_i;\pi(\cdot,z'))\leq 0, \quad i = 1,\cdots, N. \label{eq: dec-ef-macocp-2}
            \end{split}
        \end{align}
    \end{subequations}
\end{Theorem}
The proof is provided in \Cref{app: proof}.
\Cref{thm: dec-ef-macocp} enables computing $z$ \textit{without} the use of the centralized $V^l$ during execution. Specifically, each agent $i$ solves the local problem \eqref{eq: dec-ef-macocp-2} for $z_i$, which is a 1-dimensional optimization problem and can be efficiently solved using root-finding methods (e.g., \cite{chandrupatla1997new}) as in \cite{so2023solving}, then communicates $z_i$ among the other agents to obtain the maximum \eqref{eq: dec-ef-macocp-1}.
One challenge is that this maximum may not be computable if the agents are not connected.
However, in our problem setting, if one agent is not connected, it does not appear in the observations $o$ of other connected agents. Therefore, it would not contribute to the constraint-value function $V^h$ of other agents.
As a result, it is sufficient for only the connected agents to communicate their $z_i$.
Furthermore, we observe experimentally that the agents can achieve low cost while maintaining safety even if $z_i$ is not communicated (see \Cref{sec: ablation}). Thus, we do not include $z_i$ communication for our method.
The overall framework of \texttt{Def-MARL} is provided in \Cref{fig: algorithm}.

\begin{figure*}[t]
    \centering
    \includegraphics[width=.995\textwidth,trim={0 1.05cm 0 0},clip]{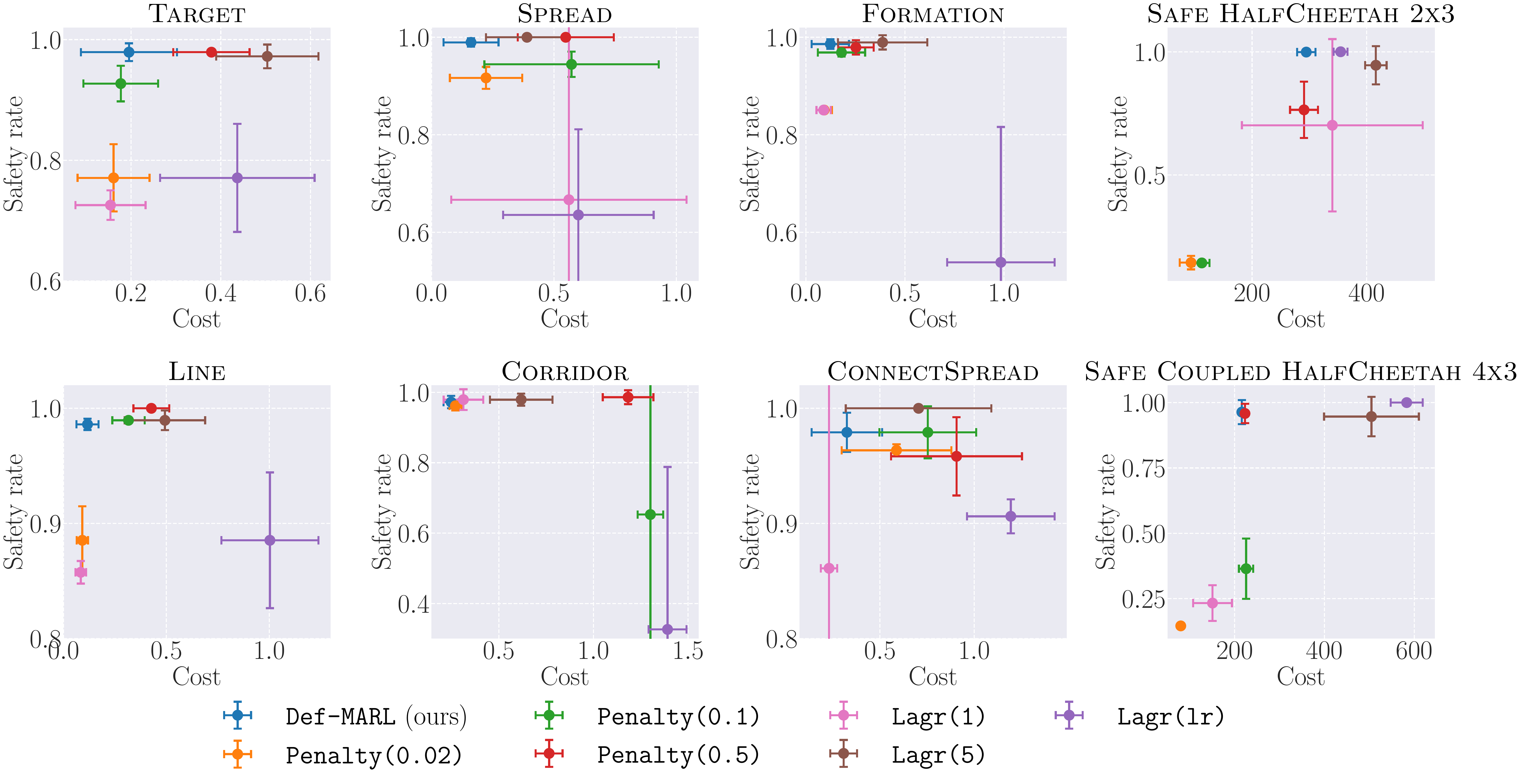}
    \caption{\textbf{Comparison on modified MPE ($N=3$) and Safe Multi-agent MuJoCo. } 
    \texttt{Def-MARL} is consistently closest to the top-left corner in all environments, achieving low cost with near $100\%$ safety rate.
The dots show the mean values and the error bars show one standard deviation.
}
    \label{fig: n=3}
\end{figure*}

\textbf{Dealing with estimation errors. }
Since there may be errors estimating $V^h$ using NN, we can reduce the resulting safety violation by modifying $h$ to add a buffer region.
Specifically, for a constant $\nu > 0$, we modify $h$ such that $h \geq \nu$ when the constraints are violated and $h \leq -\nu$ otherwise. 
We then modify \eqref{eq: dec-ef-macocp-2} to $V^h_\psi(o_i, z_i)\leq -\xi$, where $\xi\in[0, \nu]$ is a hyperparameter (where we want $\xi \approx \nu$ to emphasize more on safety). This makes $z$ more robust to estimation errors of $V^h$. We study the importance of $\xi$ in \Cref{sec: ablation}.

\section{Simulation experiments}\label{sec: experiments}

In this section, we design simulation experiments to answer the following research questions:
\begin{enumerate}[leftmargin=1.2cm,label={(\bfseries Q\arabic*):}]
\item Does \texttt{Def-MARL} satisfy the safety constraints and achieve low cost with constant hyperparameters across all environments?
\item Can \texttt{Def-MARL} achieve the global optimum of the original constrained optimization problem?
\item How stable is the training of \texttt{Def-MARL}?
\item How well does \texttt{Def-MARL} scale to larger MAS?
    \item Does the learned policy from \texttt{Def-MARL} generalize to larger MAS?
\end{enumerate}
Details for the implementation, environments, and hyperparameters are provided in \Cref{app: experiments}.

\subsection{Setup}

\textbf{Environments. }
We evaluate \texttt{Def-MARL} in two sets of simulation environments: \textit{modified} Multi-agent Particle Environments (MPE) \citep{lowe2017multi}, and Safe Multi-agent MuJoCo environments \citep{gu2023safe} (see \Cref{fig: env}). In MPE, the agents are assumed to have double integrator dynamics with bounded \emph{continuous} action spaces $[-1,1]^2$. We provide the full details of all tasks in \Cref{app: experiments}.
To increase the difficulty of the tasks, we add $3$ static obstacles to these environments. 
For Safe Multi-agent MuJoCo environments, we consider \textsc{Safe HalfCheetah 2x3} and \textsc{Safe Coupled HalfCheetah 4x3}. The agents must collaborate to make the cheetah run as fast as possible without colliding with a moving wall in front. To design the constraint function $h$, we let $\nu = 0.5$ in all our experiments and $\xi=0.4$ when solving the outer problem.

\textbf{Baselines. }
We compare our algorithm with the state-of-the-art (SOTA) MARL algorithm InforMARL \citep{nayak2023scalable} with a constraint-penalized cost $l'(x,u) = l(x,u) + \beta \max\{h(x), 0\}$, where $\beta \in\{0.02, 0.1, 0.5\}$ is a penalty parameter, and denote this baseline as \texttt{Penalty($\beta$)}. We also consider the SOTA safe MARL algorithm MAPPO-Lagrangian \citep{gu2021multi,gu2023safe}\footnote{We omit the comparison with MACPO \citep{gu2021multi,gu2023safe} as it was shown to perform similarly to MAPPO-Lagrangian but have significantly worse time complexity and wall clock time for training.}.
In addition, because the learning rate of the Lagrangian multiplier $\lambda$ is tiny ($10^{-7}$) in the official implementation
of MAPPO-Lagrangian \citep{gu2023safe}, the value of $\lambda$ during training will be largely determined by the initial value $\lambda_0$ of $\lambda$. 
We thus consider two $\lambda_0\in\{1, 5\}$.
Moreover, to compare the training stability, we consider increasing the learning rate of $\lambda$ in MAPPO-Lagrangian to $3\times10^{-3}$.\footnote{This is the smallest learning rate for $\lambda$ that does not make MAPPO-Lagrangian ignore the safety constraint. We set $\lambda_0=0.78$ following \citep{gu2023safe}.} For a fair comparison, we reimplement MAPPO-Lagrangian using the same GNN backbone as used in \texttt{Def-MARL} and InforMARL, denoted as \texttt{Lagr($\lambda_0$)} and \texttt{Lagr(lr)} for the increased learning rate one. We run each method for the same number of update steps, which is large enough for all the methods to converge.

\begin{figure*}[t]
    \centering
    \includegraphics[width=.99\textwidth,trim={0 0 0 0.5cm},clip]{figs/env_render/env_legend_cf.pdf}
    \captionsetup[subfigure]{labelformat=empty}
    \begin{subfigure}{.195\textwidth}
        \centering
        \includegraphics[width=\columnwidth]{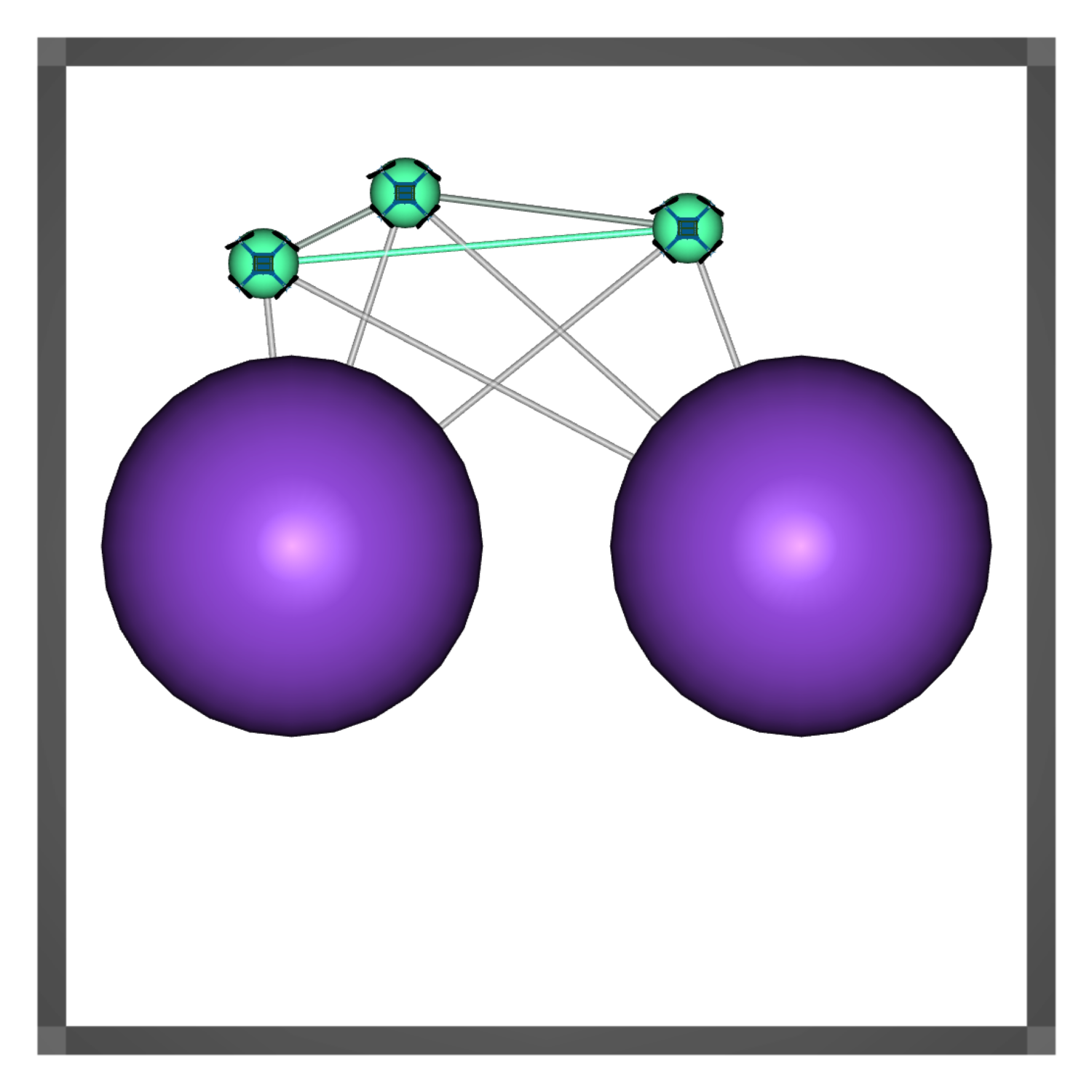}
        \caption{\texttt{Def-MARL} (ours)}
    \end{subfigure}
    \begin{subfigure}{.195\textwidth}
        \centering
        \includegraphics[width=\columnwidth]{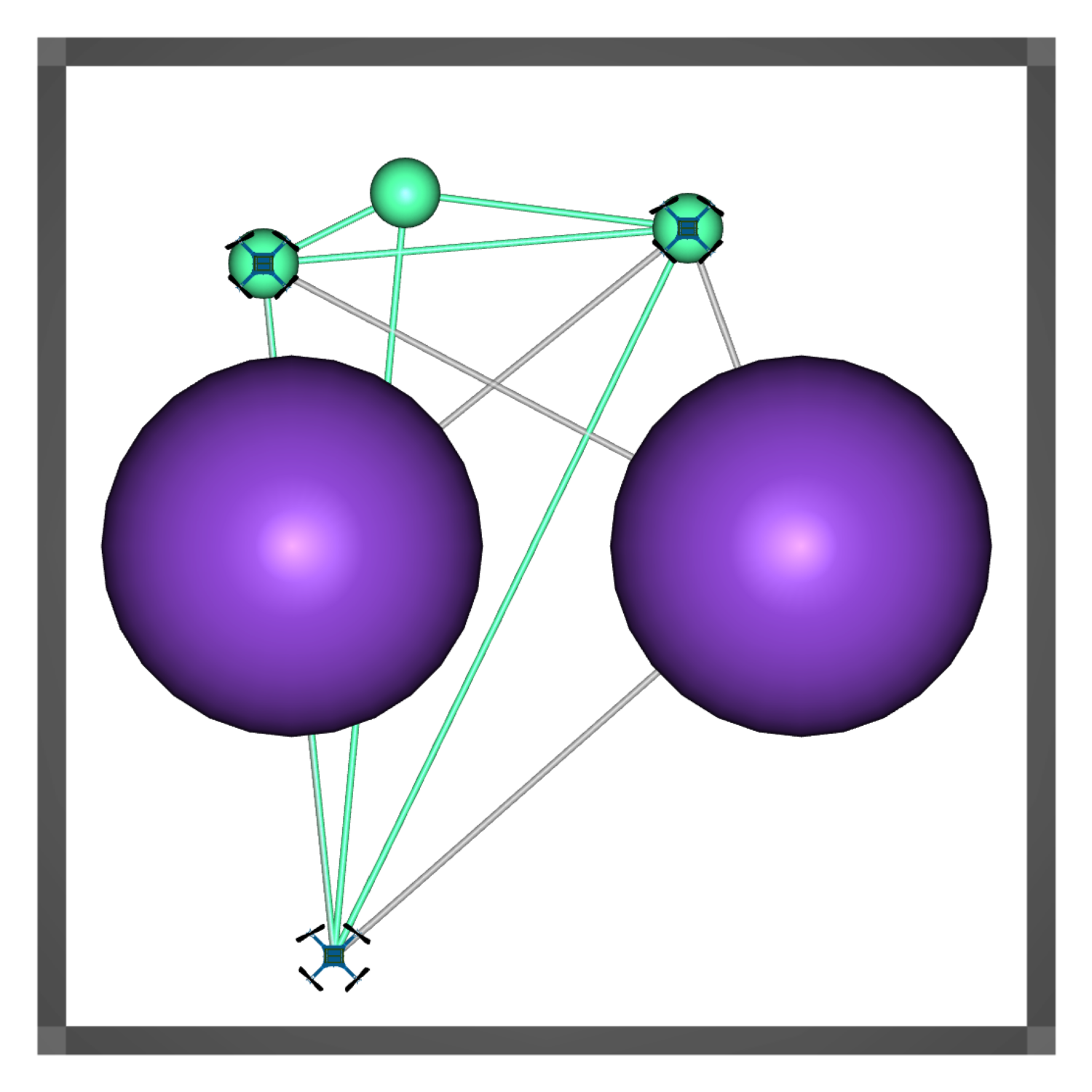}
        \caption{\texttt{Penalty(0.02)}}
    \end{subfigure}
    \begin{subfigure}{.195\textwidth}
        \centering
        \includegraphics[width=\columnwidth]{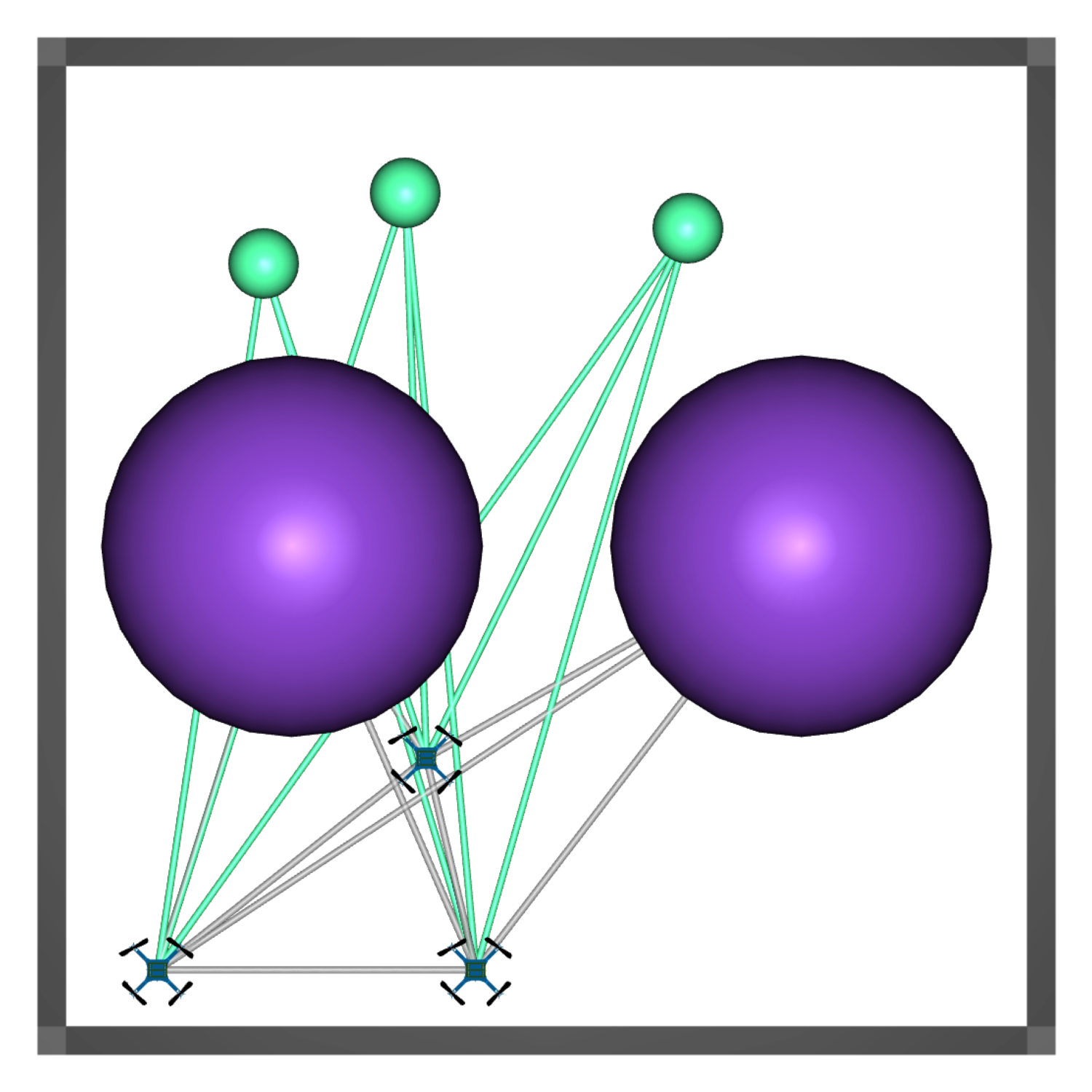}
        \caption{\texttt{Penalty(0.5)}}
    \end{subfigure}
    \begin{subfigure}{.195\textwidth}
        \centering
        \includegraphics[width=\columnwidth]{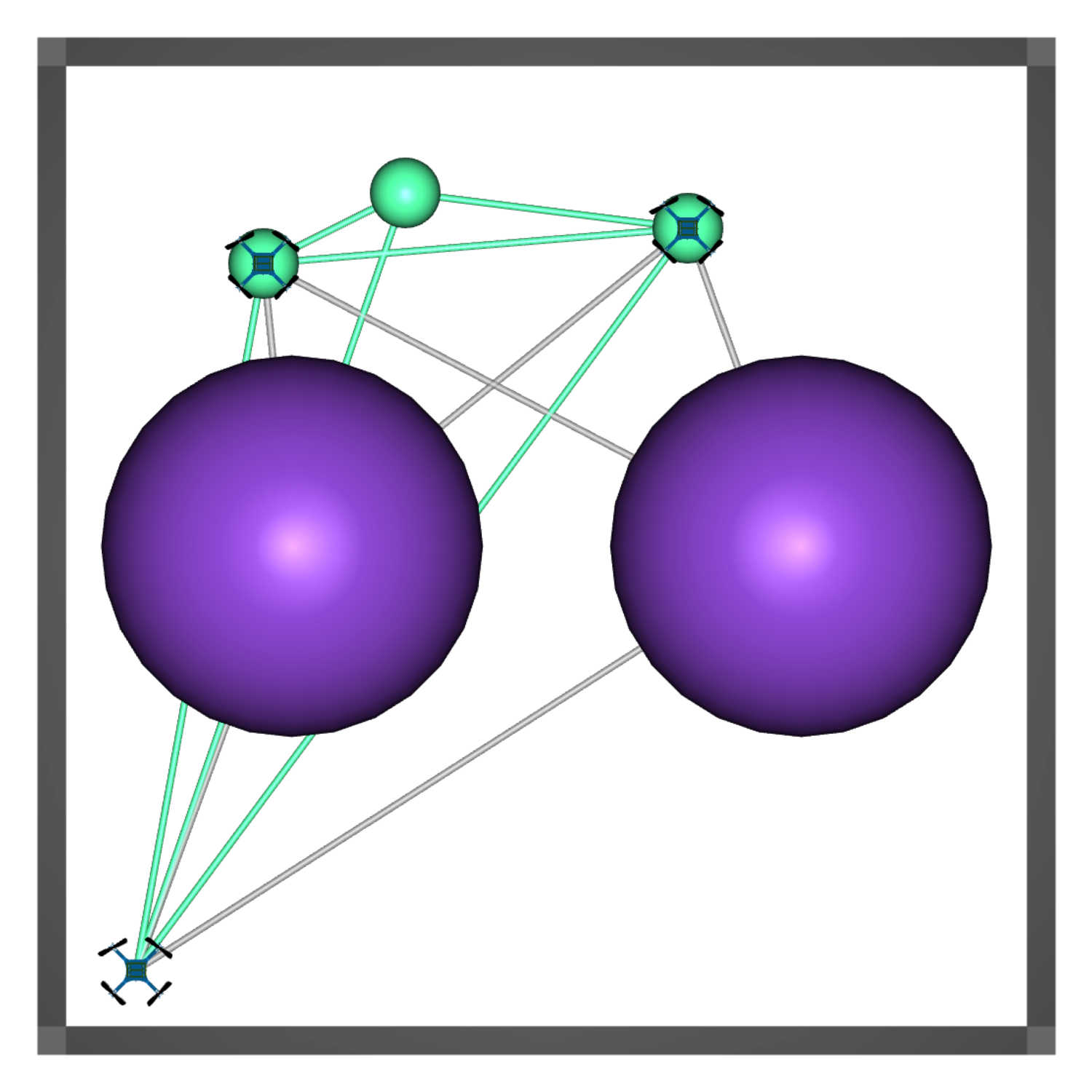}
        \caption{\texttt{Lagr(1)}}
    \end{subfigure}
    \begin{subfigure}{.195\textwidth}
        \centering
        \includegraphics[width=\columnwidth]{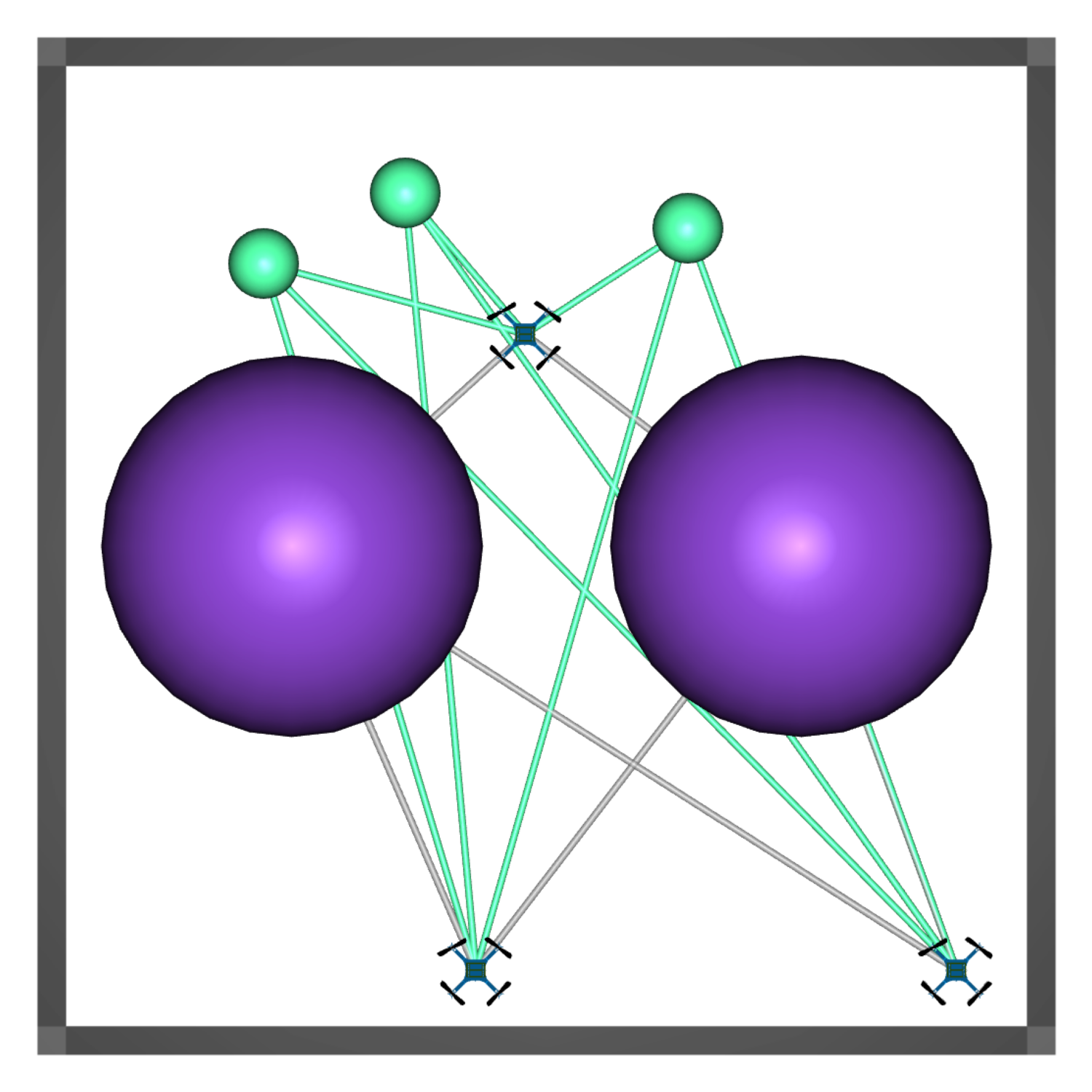}
        \caption{\texttt{Lagr(5)}}
    \end{subfigure}
    \caption{\textbf{Converged states in \textsc{Corridor}. }
\texttt{Def-MARL} achieves the global minimum, while other baselines converge to a different optimum (partly) due to training using a \textit{different} cost function.}
    \label{fig: corridor-result}
\end{figure*}

\begin{figure*}[t]
    \centering
    \includegraphics[width=0.995\textwidth,trim={0 1.1cm 0 0},clip]{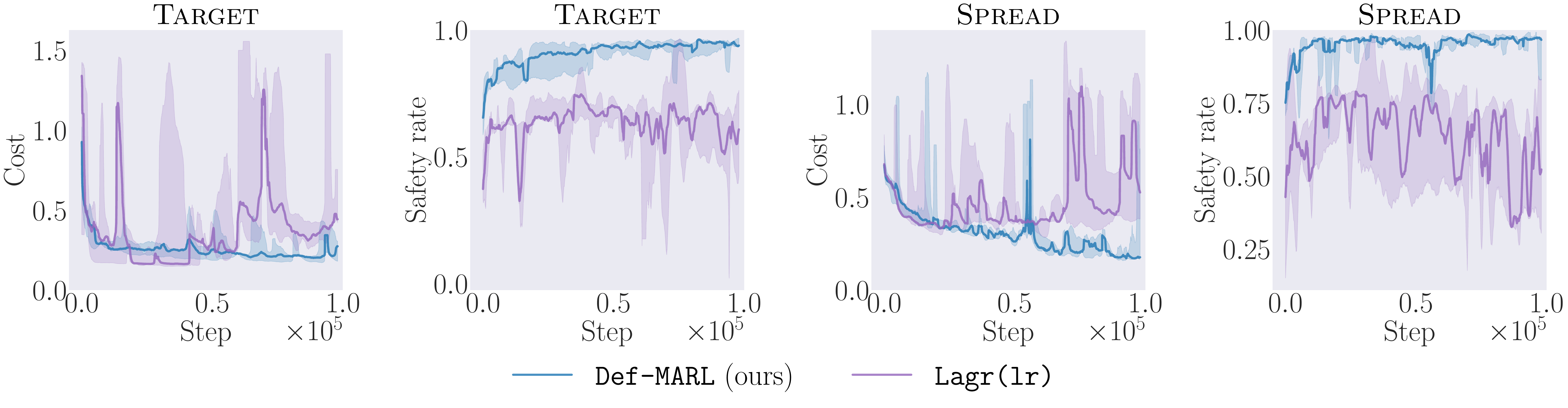}
    \caption{\textbf{Training Curves in \textsc{Target} and \textsc{Spread}. } 
    \texttt{Def-MARL} has a smoother, more stable training curve compared to \texttt{Lagr(lr)}. 
We plot the mean and shade the $\pm1$ standard deviation.}
    \label{fig: training}
\end{figure*}

\textbf{Evaluation criteria. }
Following the objective of MASOCP, we use the cost and safety rate as the evaluation criteria for the performance of all algorithms. The \textit{cost} is the cumulative cost over the trajectory $\sum_{k=0}^{T}l(x^k,u^k)$. The \textit{safety rate} is defined as the ratio of agents that remain safe over the entire trajectory, i.e., $h_i(o^k_i)\leq0, \forall k$, over all agents. Unlike the CMDP setting, we do not report the mean of constraint violations over time but the violation of the hard safety constraints. 

\begin{figure*}[t]
    \centering
    \includegraphics[width=.995\textwidth]{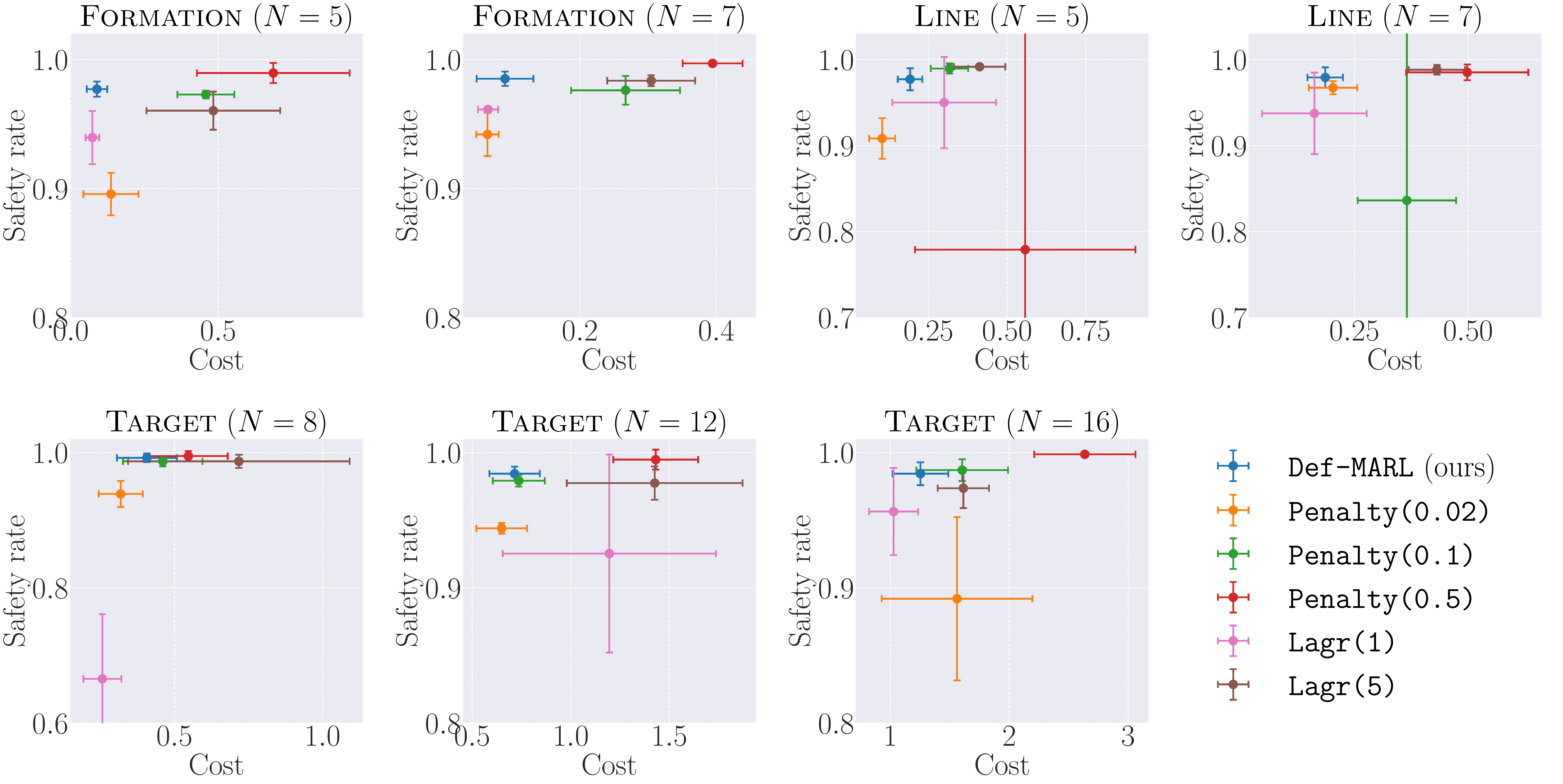}
    \caption{
    \textbf{Comparison on larger-scale modified MPE.} \texttt{Def-MARL} remains in the top-left corner even when the number of agents increases.
The dots show the mean and the error bars show one standard deviation.}
    \label{fig: n=5/7}
\end{figure*}

\subsection{Results}\label{sec: results}

We train all algorithms with $3$ different random seeds and test the converged policies on $32$ different initial conditions.
As discussed in \Cref{sec: decentralized-exe}, we \textit{disable} the communication of $z_i$ between agents (investigated in \Cref{sec: ablation}).
We draw the following conclusions.

\textbf{(Q1): \texttt{Def-MARL} achieves the best performance with constant hyperparameters across all environments. }
First, we plot the safety rate (y-axis) and cumulative cost (x-axis) for each algorithm in \Cref{fig: n=3}.
Thus, the closer an algorithm is to the top-left corner, the better it performs.
In both MPE and Safe Multi-agent MuJoCo environments, \texttt{Def-MARL} is always closest to the top-left corner, maintaining a low cost while having near $100\%$ safety rate.
For the baselines \texttt{Penalty} and \texttt{Lagr}, their performance and safety are highly sensitive to their hyperparameters.
While \texttt{Penalty} with $\beta = 0.02$ and \texttt{Lagr} with $\lambda_0 = 1$ generally have low costs, they also have frequent constraint violations.
With $\beta = 0.5$ or $\lambda_0 = 5$, they prioritize safety but at the cost of high cumulative costs.
\texttt{Def-MARL}, however, maintains a safety rate similar to the most conservative baselines (\texttt{Penalty(0.5)} and \texttt{Lagr(5)}) but has much lower costs.
We point out that no \textit{single} baseline method behaves considerably better on \textit{all} the environments: the performance of the baseline methods varies wildly between environments, demonstrating the sensitivity of these algorithms to the choice of hyperparameters.
On the contrary, \texttt{Def-MARL}, performs best in \textit{all} environments, using a single set of constant hyperparameters, which demonstrates its insensitivity to the choice of hyperparameters.

\textbf{(Q2): \texttt{Def-MARL} is able to reach the global optimum of the original problem. }
An important observation is that for \texttt{Penalty} and \texttt{Lagr} with a non-optimal $\lambda$, the cost function optimized in their training process is \emph{different} from the original cost function.
Consequently, they can have different optimal solutions compared to the original problem. Therefore, even if their training converges, they may not reach the optimal solution to the original problem.
In \Cref{fig: corridor-result}, the converged states of \texttt{Def-MARL} and four baselines are shown.
\texttt{Def-MARL} reaches the original problem's global optimum and covers all three goals.
On the contrary, the optima of \texttt{Penalty(0.02)} and \texttt{Lagr(1)} are \textit{changed} by the penalty term, so they choose to leave one agent behind to have a lower safety penalty.
With an even more significant penalty, the optima of \texttt{Penalty(0.5)} and \texttt{Lagr(5)} are changed dramatically, and they forget the goal entirely and only focus on safety. 

\begin{table}[t]
    \centering
    \caption{\textbf{Policy Generalization.} Testing \texttt{Def-MARL} on \textsc{Target} with more agents after training with $N=8$ agents.}
    \label{tab: generalizability}
    \begin{tabular}{l|ccc}
        \toprule
        \# Agent & 32 & 128 & 512 \\
        \midrule
        Safety rate & $99.8\pm0.2$ & $99.6\pm0.4$ & $99.5\pm0.3$ \\
        Cost & $-0.387\pm0.029$ & $-0.408\pm0.015$ & $-0.410\pm0.009$ \\
        \bottomrule
    \end{tabular}
\end{table}

\begin{table*}[t]
    \centering
    \begin{minipage}{.65\linewidth}
        \centering
        \caption{Effect of $z_i$ communication (\Cref{sec: decentralized-exe}) in different environments.}
        \label{tab: centralized vs decentralized}
        \begin{tabular}{c|cc|cc}
            \toprule
            \multirow{2}{*}{Environment} & \multicolumn{2}{c|}{No communication $(z \gets z_i)$} & \multicolumn{2}{c}{Communication ($z = \max_i z_i$)} \\
            & Safety rate & Cost & Safety rate & Cost \\
            \midrule
            \textsc{Target} & $97.9\pm1.5$ & $0.196\pm0.108$ & $96.9\pm3.0$ & $0.214\pm0.141$\\
            \textsc{Spread} & $99.0\pm0.9$ & $0.162\pm0.144$ & $98.6\pm1.3$ & $0.171\pm0.128$ \\
            \textsc{Formation} & $98.3\pm1.0$ & $0.123\pm0.940$ & $98.3\pm1.8$ & $0.126\pm0.100$ \\
            \textsc{Line} & $98.6\pm0.5$ & $0.117\pm0.540$ & $98.3\pm0.5$ & $0.121\pm0.630$ \\
            \textsc{Corridor} & $97.9\pm1.8$ & $0.247\pm0.390$ & $98.6\pm1.9$ & $0.255\pm0.470$ \\
            \textsc{ConnectSpread} & $97.9\pm1.7$ & $0.324\pm0.187$ & $99.0\pm0.8$ & $0.339\pm0.201$ \\
            \bottomrule
        \end{tabular}
    \end{minipage}
    \hspace{0.8em}
    \begin{minipage}{.25\linewidth}
        \centering
        \caption{Effect of varying $\xi$ (\Cref{sec: decentralized-exe}) for \textsc{Line} ($N=3$) with fixed $\nu = 0.5$.}
        \label{tab: xi-ablation}
        \begin{tabular}{c|cc}
            \toprule
            $\xi$ & Safety rate & Cost \\
            \midrule
            $0.5$ & $100.0\pm0.0$ & $0.127\pm0.061$ \\
            $0.4$ & $98.6\pm0.5$ & $0.117\pm0.540$ \\
            $0.2$ & $96.5\pm0.5$ & $0.108\pm0.044$ \\
            $0.0$ & $93.4\pm0.020$ & $0.102\pm0.035$\\
            \bottomrule
    \end{tabular}
    \end{minipage}
\end{table*}

\textbf{(Q3): Training of \texttt{Def-MARL} is more stable. }
To compare the training stability of \texttt{Def-MARL} and the Lagrangian method \texttt{Lagr(lr)}, we plot their cost and safety rate during training in \Cref{fig: training}.
\texttt{Def-MARL} has a \textit{smoother} curve compared to \texttt{Lagr(lr)}, supporting our theoretical analysis in \Cref{sec: EF}.
Due to space limits, the plots for other environments and other baseline methods are provided in \Cref{app: training-curve}.

\textbf{(Q4): \texttt{Def-MARL} can scale to more agents while maintaining high performance and safety, but is limited by GPU memory due to centralized training.}
We test the limits of \texttt{Def-MARL} on the number of agents during training by
comparing all methods on $N=5,7$ with \textsc{Formation} and \textsc{Line}, and $N=8,12,16$ with \textsc{Target} (\Cref{fig: n=5/7}).
We were unable to increase $N$ further due to GPU memory limitations due to the use of \textit{centralized} training.
For this experiment, we omit \texttt{Lagr(lr)} as it has the worst performance in MPE with $N=3$.
\texttt{Def-MARL} is closest to the upper left corner in all environments, and its performance does not decrease with an increasing number of agents.

\textbf{(Q5): The trained policy from \texttt{Def-MARL} generalizes to much larger MAS. }
To test the generalization capabilities of \texttt{Def-MARL}, we test a policy trained with $N=8$ on much larger MASs with up to $N=512$ on \textsc{Target} (\Cref{tab: generalizability}) with the same agent density to avoid distribution shift.
\texttt{Def-MARL} maintains a high safety rate and low costs despite being applied on a MAS with $64$ times more agents.

\subsection{Ablation studies}\label{sec: ablation}

Here we do ablation studies on the communication of $z_i$, and study the hyperparameter sensitivity of \texttt{Def-MARL}.

\textbf{Is communicating $z_i$ necessary? }
As introduced in \Cref{sec: decentralized-exe}, theoretically, all connected agents should communicate and reach a consensus on $z = \max_i z_i$. However, we observe in \Cref{sec: results} that the agents can perform well even if agents take $z \gets z_i$ without communicating to compute the maximum.
We perform experiments on MPE ($N=3$) to understand the impact of this approximation in \Cref{tab: centralized vs decentralized} and see that using the approximation does not result in much performance difference compared to communicating $z_i$ and using the maximum.

\textbf{Varying $\xi$ in the outer problem. }
To robustify our approach against estimation errors in $V^h$, 
we solve for a $z_i$ that is slightly more conservative by
modifying \eqref{eq: dec-ef-macocp-2} to $V_\psi^h(o_i, z_i)\leq -\xi$ (\Cref{sec: decentralized-exe}).
We now perform experiments to study the effect of different choices of $\xi$ (\Cref{tab: xi-ablation}) on \textsc{Line} ($N = 3$).
The results show that higher values of $\xi$ result in higher safety rates and slightly higher costs, while the reverse is true for smaller $\xi$.
This matches our intuition that modifying \eqref{eq: dec-ef-macocp-2} can help improve constraint satisfaction when the learned $V^h$ has estimation errors.
We thus recommend choosing $\xi$ close to $\nu$. 
We also provide the sensitivity analysis on more hyperparameters in \Cref{app: experiments}.

\section{Hardware experiments} \label{sec: hardware}

\begin{figure*}[t!]
    \centering
    \begin{subfigure}[t]{0.5\linewidth}
        \centering
        \includegraphics[width=0.99\linewidth]{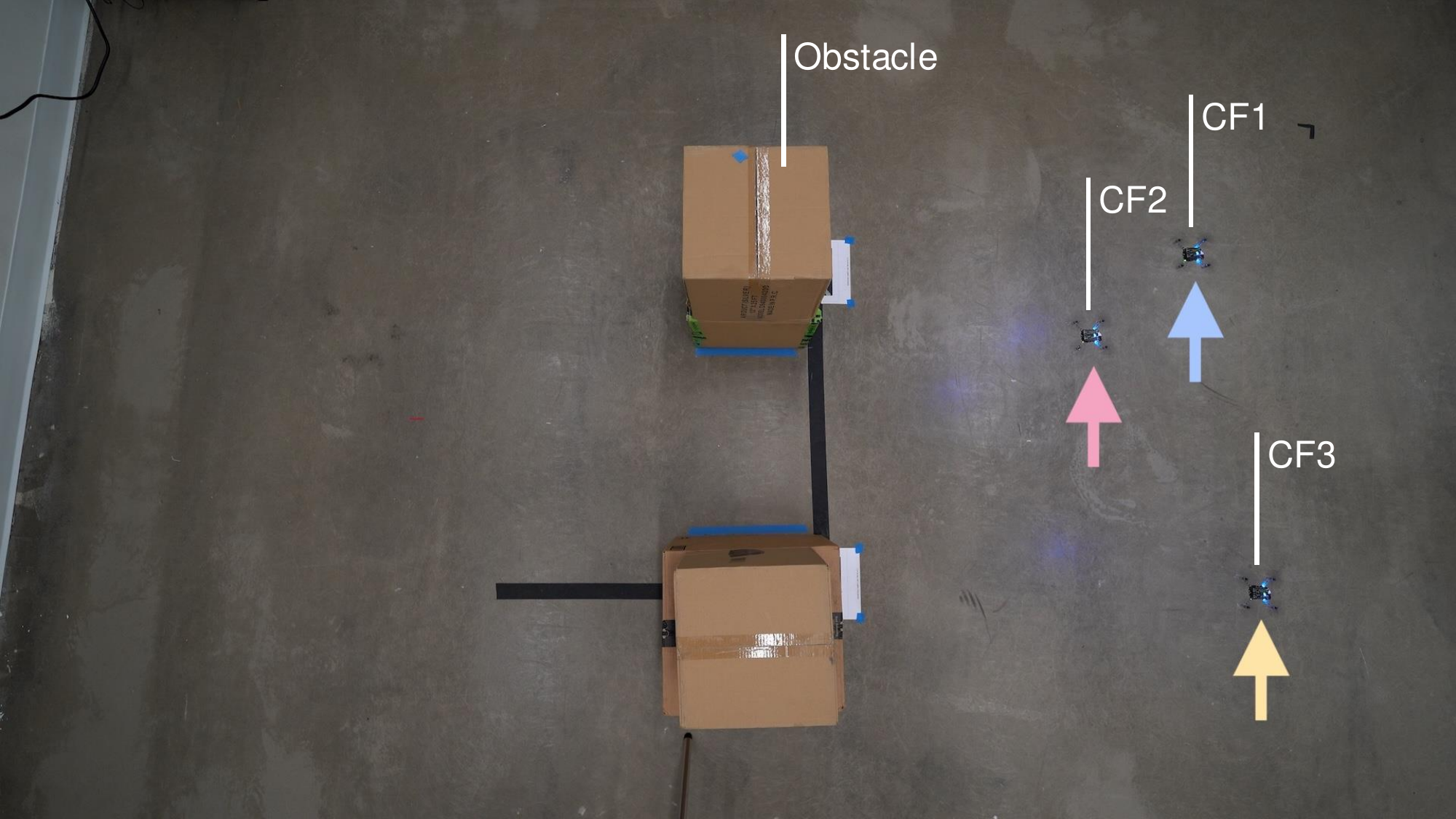}
        \caption{\textsc{Corridor}}
    \end{subfigure}\begin{subfigure}[t]{0.5\linewidth}
        \centering
        \includegraphics[width=0.99\linewidth]{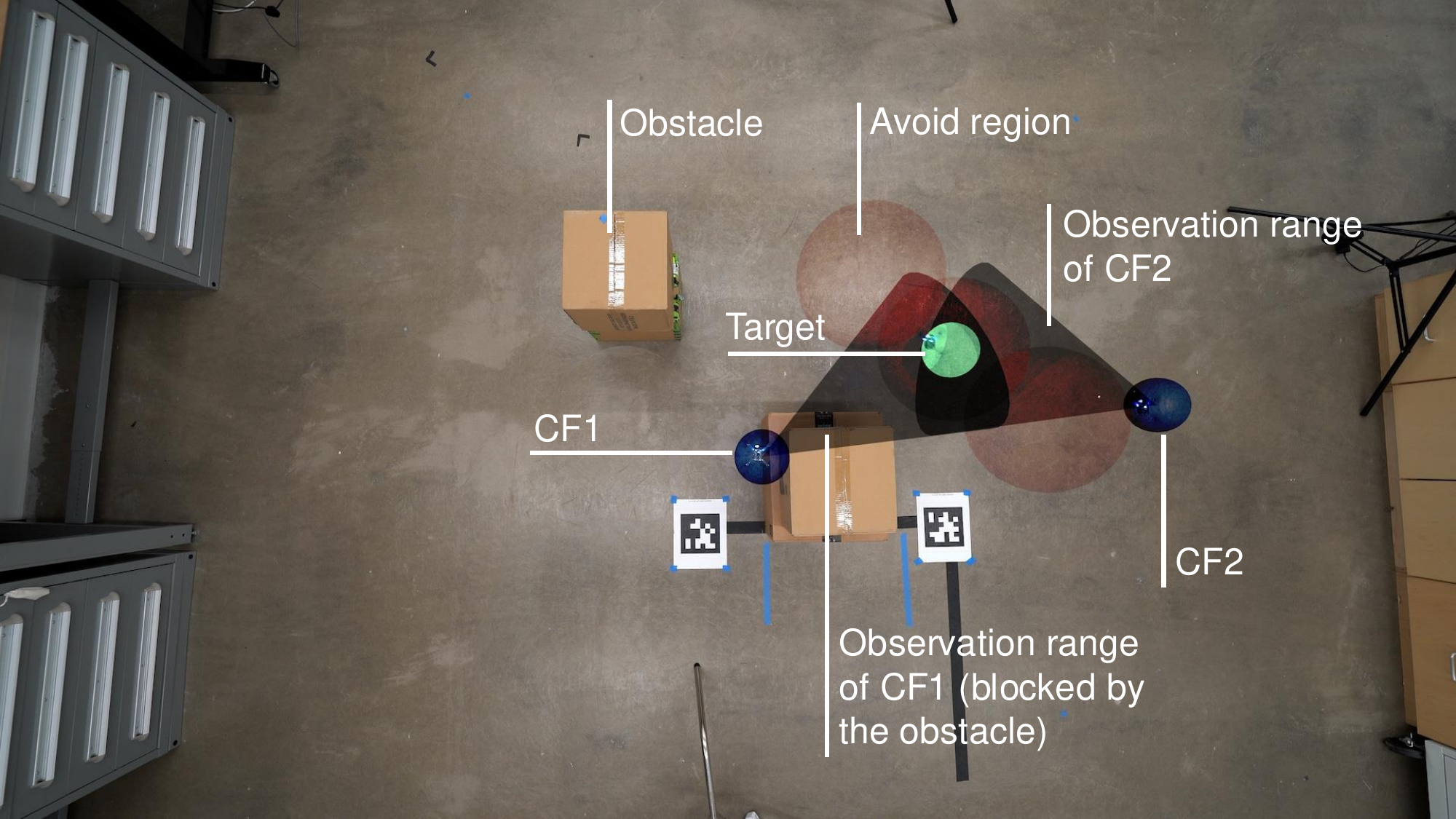}
        \caption{\textsc{Inspect}}
    \end{subfigure}
    \caption{\textbf{Hardware tasks.} We perform hardware experiments using a swarm of CF drones on the \textsc{Corridor} and \textsc{Inspect} tasks.
    In \textsc{Corridor}, the team must cross a narrow corridor and cover a set of goals collectively without prior assignment.
    In \textsc{Inspect}, the team must maintain visual contact with the target drone while staying out of the avoid zone around the target.
    }
    \label{fig: hw_task_overview}
\end{figure*}

Finally, we conduct hardware experiments on a swarm of Crazyflie (CF) drones \cite{giernacki2017crazyflie} to demonstrate \texttt{Def-MARL}'s ability to safely coordinate agents to complete complex collaborative tasks in the real world.
We consider the following two tasks.
\begin{itemize}
    \item \textsc{Corridor.} A swarm of drones collaboratively gets through a narrow corridor and reaches a set of goals without explicitly assigning drones to goals.
\item \textsc{Inspect.} Two drones collaborate to maintain direct visual contact with a target drone that follows a path shaped like an eight while staying out of the target drone's avoid zone. Visual contact only occurs when the line of sight to the target drone is not blocked by obstacles.
\end{itemize}
For both tasks, all drones have collision constraints with other drones and static obstacles.
We visualize the tasks in \Cref{fig: hw_task_overview}.

\begin{figure*}[t]
    \centering
    \includegraphics[width=0.99\linewidth]{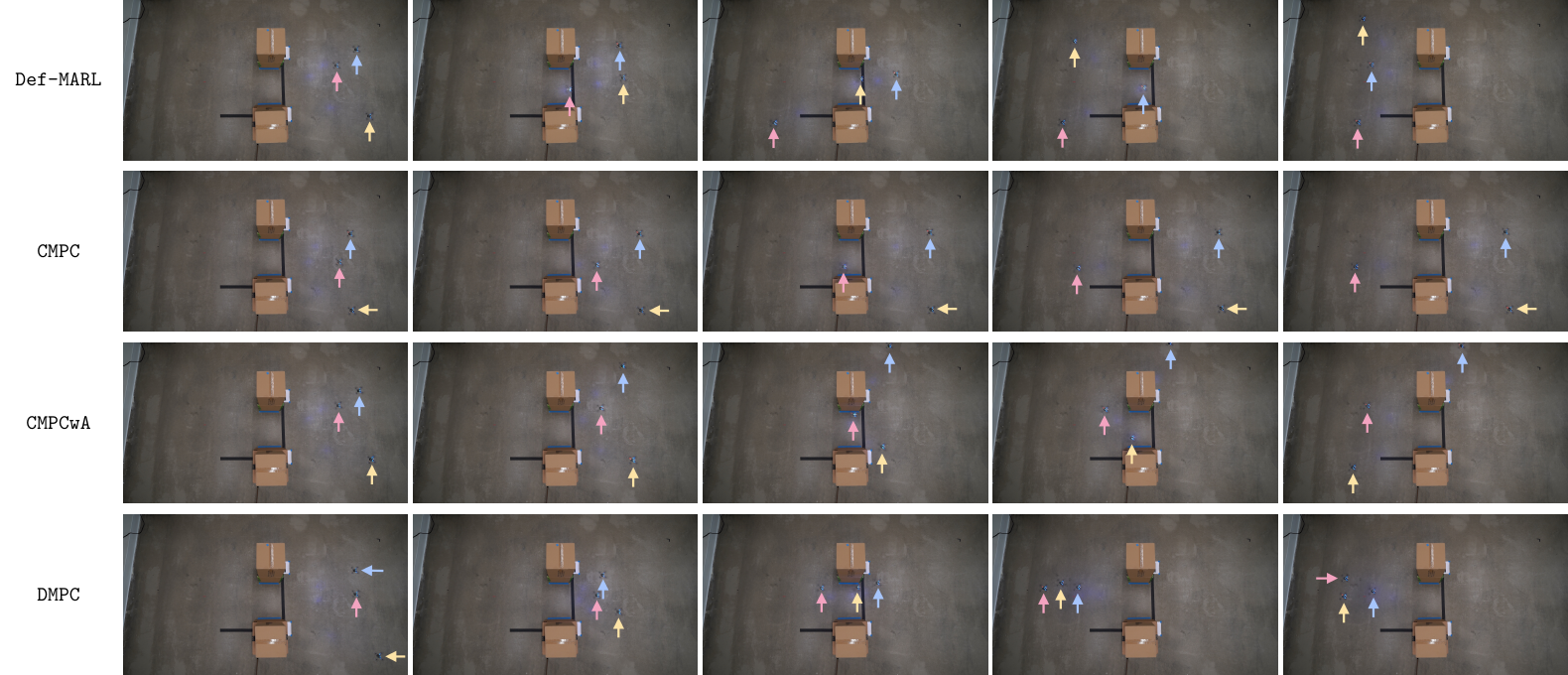}
    \caption{\textbf{Hardware Results on \textsc{Corridor} ($N=3$).} Left to right: key frames of the trajectories generated by different algorithms. Arrows with different colors indicate the positions of different drones. \texttt{Def-MARL} (top) finishes the task with $100\%$ success rate because the drones learn to cross the corridor one by one. \texttt{CMPC} and \texttt{CMPCwA} (middle) sometimes get stuck in local minima and cannot finish the task because of the highly nonconvex cost function. \texttt{DMPC} (bottom) has unsafe cases where the agents cross the unsafe radius so they cannot reach the goals because the MPC problem becomes infeasible.}
    \label{fig: hw:corridor_result_3}
\end{figure*}

\textbf{Baselines. }
\texttt{Def-MARL} has demonstrated superior performance among RL methods in simulation experiments.
Consequently, we do not consider RL methods as baselines for the hardware experiments.
Instead, we present a comparative analysis between \texttt{Def-MARL} and model predictive control (MPC), a widely employed technique in practical robotic applications.
Notably, MPC necessitates model dynamics knowledge, whereas \texttt{Def-MARL} does not.
We compare \texttt{Def-MARL} against the following two MPC baselines.
\begin{itemize}
    \item \textit{Decentralized.} We consider a decentralized MPC method (\texttt{DMPC}), where each drone tries to individually minimize the total cost and prevents collision with other controlled drones by assuming a constant velocity motion model using the current measurement of their velocity.
\item \textit{Centralized.} We also test against a centralized MPC (\texttt{CMPC}) method to better disentangle phenomena related to numerical nonlinear optimization and performing decentralized control. This method uses the same cost function used by \texttt{Def-MARL}.
\end{itemize}
Both MPC methods are implemented in CasADi \cite{andersson2019casadi} with the SNOPT nonlinear optimizer \cite{gill2005snopt}. Details for the hardware setup and experiment videos are provided in \Cref{app: hardware}.

\begin{figure*}[t]
    \centering
    \includegraphics[width=0.99\linewidth]{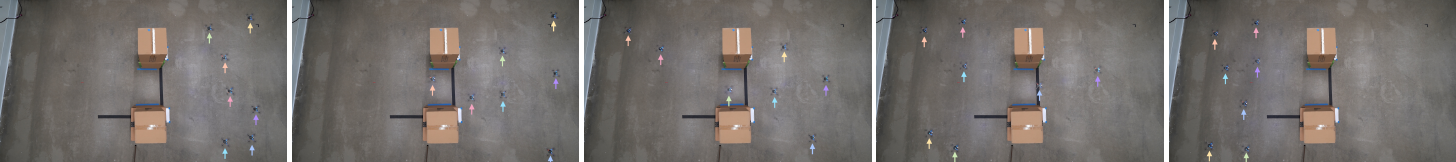}
    \caption{\textbf{Hardware Results of \texttt{Def-MARL} on \textsc{Corridor} ($N=7$).} Even in this crowded environment, \texttt{Def-MARL} maintains a success rate of $100\%$.} 
    \label{fig: hw:corridor_result_7}
\end{figure*}

\begin{figure*}[t]
    \centering
    \includegraphics[width=0.99\linewidth]{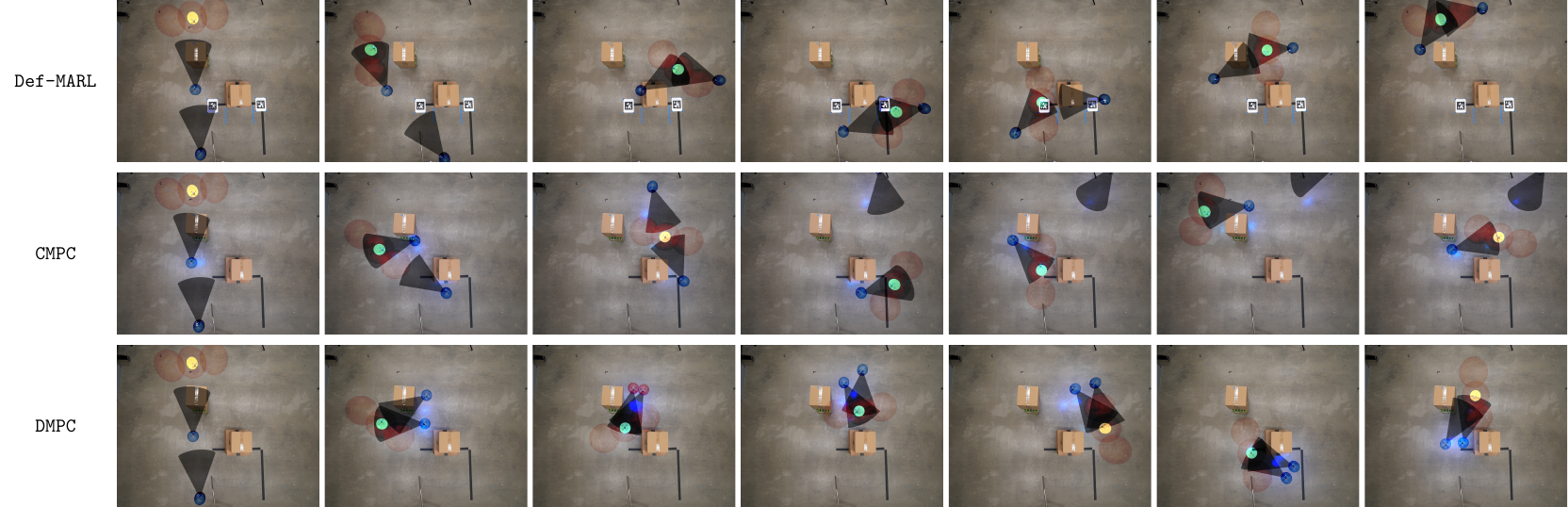}
    \caption{\textbf{Hardware Results on \textsc{Inspect}.} The CF drone overlayed with the \textcolor{teaserYellow}{yellow}/\textcolor{teaserGreen}{green} sphere is the target. The sphere turns \textcolor{teaserGreen}{green} when the target is observed and \textcolor{teaserYellow}{yellow} otherwise. The CF drones overlayed with \textcolor{teaserBlue}{blue} spheres are agents, which turn \textcolor{teaserRed}{red} if the agents become unsafe. The \textcolor{teaserObsRed}{red} spheres around the target show the avoid zone that agents cannot enter.
    \texttt{Def-MARL} finishes the task with safe and collaborative behaviors. For example, they learn to wait on two sides of the obstacles and take turns to observe the target. \texttt{CMPC} gets stuck in local minima and only moves the closest drone to the target, leaving the other drone stationary. \texttt{DMPC} makes both agents chase after the target without collaboration, and even has unsafe cases.}
    \label{fig: hw:inspect_result}
\end{figure*}

\subsection{\textsc{Corridor}} \label{sec:hw:corridor}
We run each algorithm from $16$ random initial conditions and use the task \textit{success rate} to measure their performance.
The task is defined to be successful if all the goals are covered by the agents and all agents stay safe during the whole task.
Accordingly, the success rate is defined as the number of successful tasks divided by $16$. In our tests, the success rates of \texttt{Def-MARL}, \texttt{CMPC}, and \texttt{DMPC} are $100\%$, $0\%$, and $62.5\%$. To analyze this result, we visualize the trajectory of \texttt{Def-MARL} and some failure cases of the baselines in \Cref{fig: hw:corridor_result_3}.

\textbf{\texttt{CMPC} is prone to local minima. } 
We first compare \texttt{Def-MARL} with \texttt{CMPC}.
Since we sum the distance from each goal to the closest drone, the cost function in this task is very nonconvex.
Consequently, \texttt{CMPC} results in very suboptimal solutions, where only the closest drone to the goals attempts to reach the goals, with the remaining drones left on the other side. To alleviate this issue,  although the original task does not have an explicit goal allocation, we provide a handicap to the \texttt{CMPC} methods and rerun the experiments with explicitly assigned goals for each drone. 
This simplifies the optimization problem by removing the discrete nature of goal assignment.
We name this baseline \texttt{CMPCwA} (CMPC with assignment). However, even with explicit goal assignments, we still see that sometimes one of the drones in the team gets stuck behind the corridor, resulting in a success rate of $87.5\%$. In contrast, \texttt{Def-MARL} does not succumb to this local minimum and completes the task with a success rate of $100\%$.

\textbf{\texttt{DMPC} has unsafe cases. }
As \texttt{DMPC} without goal assignment will also suffer from similar issues as \texttt{CMPC}, we choose to assign goals for each agent in this baseline. This results in a simpler problem, as explained above.
However, unlike \texttt{CMPCwA}, the agents using \texttt{DMPC} do not know the \textit{actual} actions of the other agents and can only make predictions based on their observations because of the decentralized nature of \texttt{DMPC}.
Therefore, collisions may occur if the other agents behave significantly differently from the predictions. 
In the \texttt{DMPC} row of \Cref{fig: hw:corridor_result_3}, the agents collide\footnote{For our safety, the safety radius of the drones is larger than their actual radius. Here, we mean they have entered each others' safety radius, although they have not collided in reality.} in the middle of the tasks, causing the MPC optimization problem to become infeasible and preventing the agents from reaching their goals.\footnote{Some MPC-based methods can solve the \textsc{Corridor} environment \citep{luis2020online, soria2021predictive} but assume pre-assigned goals. Additionally, these approaches need additional methods for collision avoidance (e.g., Buffered Voronoi Cells \citep{zhou2017fast}, on-demand collision avoidance approaches \citep{luis2019trajectory}), which require more domain knowledge.}

\textbf{\texttt{Def-MARL} can scale up to $7$ agents.}
We also test the scalability of \texttt{Def-MARL} with $7$ CF drones in the same environment. Notably, the size of the environment remains unchanged, so the environment becomes much more crowded and thus more challenging. We test \texttt{Def-MARL} with $9$ different random initial conditions, and it maintains a success rate of $100\%$. We visualize one of the trajectories in \Cref{fig: hw:corridor_result_7}.
Note that we were limited to only $7$ drones here due to only having $7$ drones available. However, given the simulation results, we are hopeful that \texttt{Def-MARL} can scale to larger hardware swarms.

\subsection{\textsc{Inspect}} \label{sec:hw:inspect}

We also run each algorithm from $16$ different random initial conditions.
Note that the agents may not be able to observe the target at their initial positions at the first step.
In this environment, the team of two drones should maximize the duration where the goal is observed by at least one drone.
Measuring the task performance by the number of timesteps where the target is not visible,
we obtain that the performance of \texttt{Def-MARL}, \texttt{CMPC}, and \texttt{DMPC} are $85.5\pm42.9$, $206\pm53.2$, and $251\pm59.1$, respectively.
We also report the \textit{safety rate}, defined as the ratio of tasks where all agents stay safe,
which are $100\%$, $100\%$, and $43.75\%$, respectively.
We visualize the trajectories of different methods in \Cref{fig: hw:inspect_result}.

\textbf{Agents using \texttt{Def-MARL} have better collaboration. }
The \textsc{Inspect} is designed such that collaboration between the agents is necessary to observe the target without any downtime.
One agent cannot observe the target all the time on its own because the agent's observation can be blocked by obstacles.
It also cannot simply follow the target because of the avoid region.
Using \texttt{Def-MARL}, the agents learn collaborative behaviors such as waiting on each side of the obstacles and taking turns to observe the target when the target is on their side.
The MPC methods, however, do not have such \textit{global} optimal behavior but get stuck in local minima.
For example, \texttt{CMPC} only moves the closest drone to the target, leaving the other drone stationary, while \texttt{DMPC} makes both agents chase after the target without collaboration.
Therefore, both MPC methods have small periods of time where neither drone has visual contact with the target.
In addition, similar to in \textsc{Corridor}, we observe that \texttt{DMPC} sometimes results in collisions due to the lack of coordination between drones.

\section{Conclusion}\label{sec: conclusion}

To construct safe distributed policies for real-world multi-agent systems, this paper introduces \texttt{Def-MARL} for the multi-agent safe optimal control problem, which defines safety as zero constraint violation.
\texttt{Def-MARL} takes advantage of the epigraph form of the original problem to address the training instability of Lagrangian methods in the zero-constraint violation setting.
We provide a theoretical result showing that the centralized epigraph form can be solved in a distributed fashion by each agent, which enables distributed execution of \texttt{Def-MARL}. 
Simulation results on MPE and the Safe Multi-agent MuJoCo environments suggest that, unlike baseline methods, \texttt{Def-MARL} uses a constant set of hyperparameters across all environments,
and achieves a safety rate similar to the most conservative baseline and similar performance to the baselines that prioritize performance but violate safety constraints. 
Hardware results on the Crazyflie drones demonstrate \texttt{Def-MARL}'s ability to solve complex collaborative tasks safely in the real world.

\section{Limitations}
The theoretical analysis in \Cref{sec: decentralized-exe} suggests that the connected agents must communicate $z$ and reach a consensus. If the communication on $z$ is disabled, although our experiments show that the agents still perform similarly, the theoretical optimality guarantee may not be valid. In addition, the framework does not consider noise, disturbances in the dynamics, or communication delays between agents. Finally, as a safe RL method, although safety can be theoretically guaranteed under the optimal value function and policy, this does not hold under inexact minimization of the losses. We leave tackling these issues as future work.

\section*{Acknowledgments}

This work was partly supported by the Under Secretary of Defense for Research and Engineering under Air Force Contract No. FA8702-15-D-0001. In addition, Zhang, So, and Fan are supported by the MIT-DSTA program. 
Any opinions, findings, conclusions, or recommendations expressed in this publication are those of the authors and don’t necessarily reflect the views of the sponsors.

© 2025 Massachusetts Institute of Technology.

Delivered to the U.S. Government with Unlimited Rights, as defined in DFARS Part 252.227-7013 or 7014 (Feb 2014). Notwithstanding any copyright notice, U.S. Government rights in this work are defined by DFARS 252.227-7013 or DFARS 252.227-7014 as detailed above. Use of this work other than as specifically authorized by the U.S. Government may violate any copyrights that exist in this work.

\bibliographystyle{plainnat}
\bibliography{main}

\newpage
\onecolumn
\begin{appendices}

\setcounter{Theorem}{0}
\setcounter{Lemma}{0}

\section{Proof of Proposition 1}\label{app: proof-dp}

\begin{mdframed}[style=ThmFrame]
\begin{proof}
    Under the dynamics $x^{k+1} = f(x^k,\pi(x^k))$, we have
    \begin{equation}
    \begin{aligned}
        V(x^k,z^k;\pi) & = \max\left\{\max_{p\geq k} h(x^p), \sum_{p\geq k} l(x^p,\pi(x^p)) - z^k \right\} \\
        & = \max\left\{\max\{h(x^k), \max_{p\geq k+1} h(x^p)\}, \sum_{p\geq k+1}l(x^p,\pi(x^p)) + l(x^k,\pi(x^k)) - z^k\right\} \\
        & = \max\left\{\max\{h(x^k), \max_{p\geq k+1} h(x^p)\}, \sum_{p\geq k+1}l(x^p,\pi(x^p)) - {\underbrace{\color{black}\Big[ z^k - l(x^k,\pi(x^k)) \Big]}_{\coloneqq z^{k+1}}} \right\} \\
        & = \max\left\{h(x^k), \max\left\{\max_{p\geq k+1} h(x^p), \sum_{p\geq k+1} l(x^p,\pi(x^p))-z^{k+1}\right\} \right\}\\
        & = \max\left\{h(x^k),V(x^{k+1},z^{k+1};\pi)\right\},
    \end{aligned}
    \end{equation}
    where we have defined $z^{k+1} = z^k - l(x^k,\pi(x^k))$ in the third equation. 
\end{proof}
\end{mdframed}

\section{Proof of Theorem 1}\label{app: proof}
To prove \Cref{thm: dec-ef-macocp}, first, we prove several lemmas:

{
\begin{mdframed}[style=ThmFrame]
\begin{Lemma} \label{lem: z_star_optimal}
    For any fixed state $x$, let $z^*$ denote the solution of \eqref{eq: dec-ef-macocp}, i.e.,
    \begin{subequations} \label{eq: zstar def opt}
        \begin{align}
            \min_z &\quad z, \\
            \subjto &\quad  V^h(x; \pi(\cdot, z)) \leq 0,
        \end{align}
    \end{subequations}
    and let $\pi^*$ denote $\pi(\cdot, z^*)$, i.e., it is the optimal policy for $z^*$:
    \begin{equation}
        \pi^* = \argmin_\pi V(x, z^*; \pi).
    \end{equation}
    Then, no other safe policy $\tilde{\pi}$ exists that has a strictly lower cost than $\pi^*$ while satisfying the constraints, i.e.,
    \begin{subequations} 
    \begin{align} 
        V^h(x; \tilde{\pi}) &\leq 0 \label{eq: tildepi cond 1a} \\
        V^l(x; \tilde{\pi}) &< V^l(x; \pi^*). \label{eq: tildepi cond 1b}
    \end{align}
    \end{subequations}
    In other words, $\pi^*$ is the optimal solution of the original constrained optimization problem
    \begin{subequations}
        \begin{align}
            \min_{\pi} &\quad V^l(x; \pi), \\
            \subjto &\quad  V^h(x; \pi) \leq 0.
        \end{align}
    \end{subequations}
\end{Lemma}
\end{mdframed}

Before proving this lemma, we first prove the following lemma.
\begin{mdframed}[style=ThmFrame]
\begin{Lemma} \label{lem: dagger}
    Suppose that such a $\tilde{\pi}$ exists. Then, there exists a $z^\dagger \coloneqq V^l(x; \tilde{\pi}) - V^h(x; \tilde{\pi})$ for which the optimal policy $\pi^\dagger$ for $z^\dagger$ satisfies the conditions for $\tilde{\pi}$ in \eqref{eq: tildepi cond 1a} and \eqref{eq: tildepi cond 1b}, i.e.,
    \begin{subequations}
    \begin{align}
        V^h(x; \pi^\dagger) &\leq V^h(x; \tilde{\pi}) \leq 0, \label{eq: tildepi cond 2a} \\
        V^l(x; \pi^\dagger) &\leq V^l(x; \tilde{\pi}) < V^l(x; \pi^*). \label{eq: tildepi cond 2b}
    \end{align}
    \end{subequations}
\end{Lemma}
\end{mdframed}

\begin{mdframed}[style=ProofFrame]
\begin{proof}
    {
    Since $\pi^\dagger$ is optimal for $z^\dagger$, we have that
    \begin{equation}
        V(x, z^\dagger; \pi^\dagger) \leq V(x, z^\dagger; \tilde{\pi}).
    \end{equation}
    This implies that, by definition of $z^\dagger$,
    \begin{align}
        \max\Big\{ V^h(x; \pi^\dagger),\; V^l(x; \pi^\dagger) - z^\dagger \Big\} 
        &\leq \max\Big\{ V^h(x; \tilde{\pi}),\; V^l(x; \tilde{\pi}) - z^\dagger \Big\}, \\
        &= V^h(x; \tilde{\pi}).
    \end{align}
    In particular,
    \begin{equation}
        V^h(x; \pi^\dagger) \leq V^h(x, \tilde{\pi}),
    \end{equation}
    and
    \begin{align}
        V^l(x; \pi^\dagger) - \Big( V^l(x; \tilde{\pi}) - V^h(x; \tilde{\pi}) \Big) &\leq V^h(x, \tilde{\pi}), \\
        \implies V^l(x; \pi^\dagger) &\leq V^l(x; \tilde{\pi}).
    \end{align}
    which proves \eqref{eq: tildepi cond 2a} and \eqref{eq: tildepi cond 2b}.
    }
\end{proof}
\end{mdframed}

We are now ready to prove \Cref{lem: z_star_optimal}.
\begin{mdframed}[style=ProofFrame]
\begin{proof}[Proof of \Cref{lem: z_star_optimal}]
    We prove this by contradiction.

    Suppose that such a $\tilde{\pi}$ exists. By \Cref{lem: dagger}, there exists $z^\dagger$ and $\pi^\dagger$ that satisfies the conditions for $\tilde{\pi}$ in \eqref{eq: tildepi cond 1a} and \eqref{eq: tildepi cond 1b}.
    Since $\pi^*$ is optimal for $z^*$, this implies that
    \begin{equation}
        \max\Big\{ V^h(x; \pi^*),\; V^l(x; \pi^*) - z^* \Big\} \leq \max\Big\{ V^h(x; \pi^\dagger),\; V^l(x; \pi^\dagger) - z^* \Big\}.
    \end{equation}
    We now consider two cases depending on the value of the $\max$ on the right.
    \begin{mdframed}[style=Frame]
    \textbf{Case 1 ($V^h(x; \pi^\dagger) \leq V^l(x; \pi^\dagger) - z^*$):}
    For this case, $\max\Big\{ V^h(x; \pi^\dagger),\; V^l(x; \pi^\dagger) - z^* \Big\} = V^l(x; \pi^\dagger) - z^*$. This implies that
    \begin{equation}
        V^l(x; \pi^*) - z^* \leq V^l(x; \pi^\dagger) - z^* \qquad \iff \qquad V^l(x; \pi^*) \leq V^l(x; \pi^\dagger).
    \end{equation}
    However, this contradicts our assumption that $V^l(x; \pi^\dagger) \leq V^l(x; \tilde{\pi}) < V^l(x; \pi^*)$ from \eqref{eq: tildepi cond 1b}.

    \noindent\textbf{Case 2 ($V^h(x; \pi^\dagger) > V^l(x; \pi^\dagger) - z^*$):}
    For this case, $\max\Big\{ V^h(x; \pi^\dagger),\; V^l(x; \pi^\dagger) - z^* \Big\} = V^h(x; \pi^\dagger)$. This implies that
    \begin{equation} \label{eq: Vh star leq dagger}
        V^h(x; \pi^*) \leq V^h(x; \pi^\dagger)
    \end{equation}
    and 
    \begin{equation} \label{eq: zstar equiv}
        V^l(x; \pi^*) - z^* \leq V^h(x; \pi^\dagger) \qquad \implies \qquad V^l(x; \pi^*) - V^h(x; \pi^\dagger) \leq  z^*.
    \end{equation}
    However, if we examine the definition of $z^\dagger$, we have that
    \begin{align}
        z^\dagger
        &= V^l(x; \tilde{\pi}) - V^h(x; \tilde{\pi}) \\
        &\leq V^l(x; \tilde{\pi}) - V^h(x; \pi^\dagger) && \text{(from \eqref{eq: tildepi cond 2a} and \eqref{eq: Vh star leq dagger})} \\
        &< V^l(x; \pi^*) - V^h(x; \pi^\dagger) && \text{(from \eqref{eq: tildepi cond 1b})} \\
        &\leq z^* && \text{(from \eqref{eq: zstar equiv})}.
    \end{align}
    This contradicts our definition of $z^*$ being the optimal solution of \eqref{eq: zstar def opt}, since $z^\dagger$ satisfies $V^h(x; \pi(\cdot, z^\dagger)) \leq 0$ but is also strictly smaller than $z^*$.
    \end{mdframed}
    Since both cases lead to a contradiction, no such $\tilde{\pi}$ can exist.
\end{proof}
\end{mdframed}

We also prove the following lemma.

\begin{mdframed}[style=ThmFrame]
    \begin{Lemma} \label{lem:monotone}
        For any fixed state $x$, let $z^*$ denote the solution of \eqref{eq: dec-ef-macocp}, i.e.,
        \begin{subequations} \begin{align}
                \min_z &\quad z, \\
                \subjto &\quad  V^h(x; \pi(\cdot, z)) \leq 0,
            \end{align}
        \end{subequations}
        and let $\pi_{z^*}$ denote $\pi(\cdot, z^*)$, i.e., it is the optimal policy for $z^*$:
        \begin{equation}
            \pi_{z^*} = \argmin_\pi V(x, z^*; \pi).
        \end{equation}
        Assuming that there does not exist another $z$ such that $V^l(x; \pi_z) = V^l(x; \pi_{z^*})$. Then for any $\epsilon \geq 0$,
        \begin{equation}
            V^h(x; \pi_{z^* + \epsilon}) \leq V^h(x; \pi_{z^*}).
        \end{equation}
    \end{Lemma}
\end{mdframed}
\begin{mdframed}[style=ProofFrame]
    \begin{proof}
        Since $\pi_{z^* + \epsilon}$ is optimal for $z = z^* + \epsilon$,
        \begin{equation}
            \max\{ V^h(x; \pi_{z^* + \epsilon}), V^l(x; \pi_{z^*} + \epsilon) - (z^* + \epsilon) \} \leq \max\{ V^h(x; \pi_{z^*}), V^l(x; \pi_{z^*}) - (z^* + \epsilon) \} .
        \end{equation}
        For the $\max$ on the right-hand side, if $V^h(x; \pi_{z^*}) \geq V^l(x; \pi_{z^*}) - (z^* + \epsilon)$, then we immediately obtain our desired result
        \begin{equation}
            V^h(x; \pi_{z^* + \epsilon}) \leq V^h(x; \pi_{z^*}).
        \end{equation}
        We thus suppose that $V^h(x; \pi_{z^*}) \geq V^l(x; \pi_{z^*}) - (z^* + \epsilon)$, and obtain that
        \begin{subequations}
        \begin{align}
            V^h(x; \pi_{z^* + \epsilon}) &\leq V^l(x; \pi_{z^*}) - (z^* + \epsilon), \\
            V^l(x; \pi_{z^* + \epsilon}) &\leq V^l(x; \pi_{z^*}). \label{eq: rip: Vl thing}
        \end{align}
        \end{subequations}

        Now, since $\pi_{z^*}$ is optimal for $z = z^*$,
        \begin{equation}
            \max\{ V^h(x; \pi_{z^*}), V^l(x; \pi_{z^*}) - z^* \} \leq \max\{ V^h(x; \pi_{z^* + \epsilon}), V^l(x; \pi_{z^* + \epsilon}) - z^* \} .
        \end{equation}
        We now split into two cases depending on the $\max$ on the right-hand side.
        \begin{mdframed}[style=Frame]
            \textbf{Case 1 ($V^h(x; \pi_{z^* + \epsilon}) \geq V^l(x; \pi_{z^* + \epsilon}) - z^*$):}
            This implies that
            \begin{equation}
                V^l(x; \pi_{z^*}) - z^* \leq V^h(x; \pi_{z^* + \epsilon}),
            \end{equation}
            hence,
            \begin{equation}
                V^h(x; \pi_{z^* + \epsilon}) \leq V^l(x; \pi_{z^*}) - (z^* + \epsilon) \leq V^h(x; \pi_{z^* + \epsilon}) - \epsilon,
            \end{equation}
            which is a contradiction, so this case does not occur.

            \noindent\textbf{Case 2 ($V^h(x; \pi_{z^* + \epsilon}) < V^l(x; \pi_{z^* + \epsilon}) - z^*$):}
            This implies that
            \begin{equation}
                V^l(x; \pi_z^*) - z^* \leq V^l(x; \pi_{z^* + \epsilon}) - z^*,
            \end{equation}
            hence, $V^l(x; \pi_z^*) \leq V^l(x; \pi_{z^* + \epsilon})$.
            Combining this with \eqref{eq: rip: Vl thing} gives us that $V^l(x; \pi_z^*) = V^l(x; \pi_{z^* + \epsilon})$.
            However, from our assumption, this implies that $\pi_z^* = \pi_{z^* + \epsilon}$ and thus our desired result of $V^h(x; \pi_{z^* + \epsilon}) \leq V^h(x; \pi_{z^*})$.
        \end{mdframed}
    \end{proof}
\end{mdframed}

We can now prove \Cref{thm: dec-ef-macocp}, which follows as a consequence of \Cref{lem: z_star_optimal} and \Cref{lem:monotone}.
\begin{mdframed}[style=ProofFrame]
\begin{proof}[Proof of \Cref{thm: dec-ef-macocp}]
    Since $V^h( x;\pi) = \max_i V_i^h(x_i, o_i;\pi)$, \Cref{lem: z_star_optimal} implies that \Cref{eq: ef-macocp} is equivalent to
{
    \begin{equation}
        z^* \coloneqq \min_z \{ z \mid \max_i V^h_i(x_i, o_i; \pi(\cdot, z) ) \leq 0\}.
    \end{equation}
    We now show that this is equivalent to the following distributed implementation (equal to \eqref{eq: dec-ef-macocp}):
    \begin{align}
        z_i &\coloneqq \min\{ z \mid V^h_i(x_i, o_i; \pi(\cdot, z) ) \leq 0 \} \label{eq: app:zi}, \\
        z_{\text{distr}} &= \max_i z_i.
    \end{align}
    We will now prove equality via a double inequality proof.
    \begin{mdframed}[style=Frame]
        \textbf{($z_{\text{distr}} \leq z^*$):}
        By definition of $z^*$,
        \begin{equation}
            V^h_i(x_i, o_i; \pi(\cdot, z^*) ) \leq 0, \quad \forall i.
        \end{equation}
        However, since $z_i$ \eqref{eq: app:zi} is optimal, $z_i \leq z^*$. Hence,
        \begin{equation}
            z_{\text{distr}} = \max_i z_i \leq z^*.
        \end{equation}

        \noindent \textbf{($z_{\text{distr}} \geq z^*$):}
        By definition of $z_{\text{distr}}$,
        \begin{equation}
            z_{\text{distr}} \geq z_i, \quad \forall i.
        \end{equation}
        Using \Cref{lem:monotone}, this implies that
        \begin{equation}
            V^h_i(x_i, o_i; \pi(\cdot, z_{\text{distr}}) ) \leq 0, \quad \forall i.
        \end{equation}
        Hence, $z_{\text{distr}}$ satisfies $\max_i V^h_i(x_i, o_i; \pi(\cdot, z_{\text{distr}}) ) \leq 0$. Since $z^*$ is the smallest $z$ that still satisfies this constraint,
        \begin{equation}
            z^* \leq z_{\text{distr}}.
        \end{equation}
    \end{mdframed}

    We have thus proved that $z^* = z_{\text{distr}}$, i.e., $z^*$ can be computed in a distributed fashion.
}
\end{proof}
\end{mdframed}
}

\section{Discussion on Importance of \Cref{thm: dynamic-program}} \label{app: discussion-dp}
Establishing \Cref{thm: dynamic-program} is \textit{key} to \texttt{Def-MARL}. Namely,
\begin{enumerate}[leftmargin=1cm]
    \item Satisfying dynamic programming implies that the value function is \textit{Markovian}. In other words, for a given $z^0$, the value at the $k$th timestep is \textit{only} a function of $z^k$ and $x^k$ instead of the $z^0$ and the \textit{entire} trajectory up to the $k$th timestep.
    \item Consequently, this implies that the optimal policy will also be Markovian and is \textit{only} a function of $z^k$ and $x^k$.
    \item Rephrased differently, since the value function is Markovian, this implies that, for a given $z^0$ and $x^0$, the value at the $k$th timestep is \textit{equal} to the value (at the initial timestep) of a \textit{new} problem where we start with $\tilde{z}^0 = z^k$ and $\tilde{x}^0 = x^k$.
    \item Since we relate the value function of consecutive timesteps, given a value function estimator, we can now control the bias-variance tradeoff of the value function estimate by using k-step estimates instead of the Monte Carlo estimates.
    \item Instead of only using the k-step estimates for a single choice of k, we can compute a weighted average of the k-step estimates as in GAE to further control the bias-variance tradeoff.
\end{enumerate}

\section{Algorithm Pseudocode}

We describe the centralized training process of \texttt{Def-MARL} in \Cref{alg: training} and the distributed execution process in \Cref{alg: execution}. 

\begin{algorithm}[h]
    \caption{\texttt{Def-MARL} centralized training}\label{alg: training}
    \begin{algorithmic}
    \State \textbf{Initialize:} Policy NN $\pi_\theta$, cost value function NN $V^l_\phi$, constraint value function NN $V^h_\psi$.
    \While{Training not end}
        \State Randomly sampling initial conditions $x^0$, and the initial $z^0\in[z_\mathrm{min}, z_\mathrm{max}]$.
        \State Use $\pi_\theta$ to sample trajectories $\{x^0,\dots,x^T\}$, with $z$ dynamics \eqref{eq: dynamic-program}.
        \State Calculate the cost value function $V^l_\phi(x, z)$ and the constraint value function $V^h_\psi(o_i, z)$.
        \State Calculate GAE with the total value function \eqref{eq: V-def}. 
        \State Update the value functions $V^l_\phi$ and $V^h_\psi$ using TD error.
        \State Update the $z$-conditioned policy $\pi_\theta(\cdot, z)$ using PPO loss. 
    \EndWhile
    \end{algorithmic}
\end{algorithm}

\begin{algorithm}[h]
    \caption{\texttt{Def-MARL} distributed execution}\label{alg: execution}
    \begin{algorithmic}
    \State \textbf{Input:} Learned policy NN $\pi_\theta$, constraint value function NN $V^h_\psi$.
    \For{$k = 0, \dots, T$}
        \State Get $z_i$ for each agent by solving the distributed EF-MASOCP outer problem \eqref{eq: dec-ef-macocp-2}.
        \If{$z$ communication enabled}
            \State The connected agents $j$ communicate $z_j$ and reach a consensus $z = \max_j z_j$.
            \State Set $z_i = z$ for all agents in the connected graph.
        \EndIf
        \State Get decentralized policy $\pi_i(\cdot) = \pi_\theta(\cdot, z_i)$.
        \State Execute control $u^k_i = \pi_i(o^k_i)$.
    \EndFor
    \end{algorithmic}
\end{algorithm}

\section{Simulation experiments}\label{app: experiments}

\subsection{Computation resources}

The experiments are run on a 13th Gen Intel(R) Core(TM) i7-13700KF CPU with 64GB RAM and an NVIDIA GeForce RTX 3090 GPU. The training time is around $6$ hours ($10^5$ steps) for \texttt{Def-MARL} and \texttt{Lagr}, and around $5$ hours for \texttt{Penalty}. 

\subsection{Environments}\label{app: experiments-environments}

\subsubsection{Multi-partical environments (MPE)}

We use directed graphs $\mathcal G = (\mathcal V, \mathcal E)$ to represent MPE, where $\mathcal V$ is the set of nodes containing the objects in the multi-agent environment (e.g., agents $\mathcal V_a$, goals $\mathcal V_g$, landmarks $\mathcal V_l$, and obstacles $\mathcal V_o$). $\mathcal E\subseteq \{(i, j)\;|\;i\in \mathcal V_a, j\in \mathcal V\}$ is the set of edges, denoting the information flow from a sender node $j$ to a receiver agent $i$. An edge $(i, j)$ exists only if the communication between node $i$ and $j$ can happen, which means the distance between node $i$ and $j$ should be within the communication radius $R$ in partially observable environments. We define the neighborhood of agent $i$ as $\mathcal N_i\coloneqq\{j\;|\;(i,j)\in\mathcal E\}$. The node feature $v_i$ includes the states of the node $x_i$ and a one-hot encoding of the type of the node $i$ (e.g., agent, goal, landmark, or obstacle), e.g., $[0, 0, 1]^\top$ for agent nodes, $[0, 1, 0]^\top$ for goal nodes, and $[1, 0, 0]^\top$ for obstacle nodes. The edge feature $e_{ij}$ includes the information passed between the sender node $j$ and the receiver node $i$ (e.g., relative positions and velocities).

We consider $6$ MPE: \textsc{Target}, \textsc{Spread}, \textsc{Formation}, \textsc{Line}, \textsc{Corridor}, and \textsc{ConnectSpread}. In each environment, the agents need to work collaboratively to finish some tasks:

\begin{itemize}
    \item \textsc{Target} \citep{nayak2023scalable}: Each agent tries to reach its preassigned goal. 
    \item \textsc{Spread} \citep{dames2017detecting}: The agents are given a set of (not preassigned) goals to cover.
    \item \textsc{Formation} \citep{agarwal2020learning}: Given a landmark, the agents should spread evenly on a circle with the landmark as the center and a given radius. 
    \item \textsc{Line} \citep{agarwal2020learning}: Given two landmarks, the agents should spread evenly on the line between the landmarks. 
    \item \textsc{Corridor}: A set of agents and goals are separated by a narrow corridor, whose width is smaller than $4r_a$ where $r_a$ is the radius of agents. The agents should go through the corridor and cover the goals. 
    \item \textsc{ConnectSpread}: A set of agents and goals are separated by a large obstacle with a diameter larger than the communication radius $R$. The agents should cover the goals without colliding with obstacles or each other while also maintaining the connectivity of all agents.
\end{itemize}

We consider $N=3$ agents for all environments and $N=5$ and $7$ agents in the \textsc{Formation} and \textsc{Line} environments. To make the environments more difficult than the original ones \citep{nayak2023scalable}, we add $3$ static obstacles in the first $4$ environments. 

In our modified MPE, the state of the agent $i$ is given by $x_i = [p^x_i, p^y_i, v^x_i, v^y_i]^\top$, where $[p^x_i,p^y_i]^\top \coloneqq p\in\mathbb R^2$ is the position of agent $i$, and $[v^x_i, v^y_i]$ is the velocity. The control inputs are given by $u_i = [a^x_i, a^y_i]^\top$, i.e., the acceleration along each axis. The joint state is defined by concatenation: $x = [x_1;\dots;x_N]$. The agents are modeled as double integrators with dynamics
\begin{equation}
    \dot x_i = \begin{bmatrix}
        v^x_i & v^y_i & a^x_i & a^y_i
    \end{bmatrix}^\top.
\end{equation}
The agents' control inputs are limited by $[-1, 1]$, and the velocities are limited by $[-1, 1]$. The agents have a radius $r_a = 0.05$, and the communication radius is assumed to be $R = 0.5$. The area side length $L$ is $1.0$ for the \textsc{Corridor} and the \textsc{ConnectSpread} environments and $1.5$ for other environments. The radius of the obstacles $r_o$ is $0.4$ in the \textsc{Corridor} environment, $0.25$ in the \textsc{ConnectSpread} environment, and $0.05$ for other environments. All environments use a simulation time step of $0.03$s and a total horizon of $T=128$.

The observation of the agents $o_i$ includes the node features of itself, its neighbors $j\in\mathcal N_i$, and the edge features of the edge connecting agent $i$ and its neighbors. The node features include neighbors' states $x_j$ and its type ($[0, 0, 1]$ for agents, $[0, 1, 0]$ for goals and landmarks, and $[1, 0, 0]^\top$ for the obstacles). The edge features are the relative states $e_{ij} = x_i - x_j$. 

\begin{figure}[t]
    \centering
    \includegraphics[width=0.3\columnwidth]{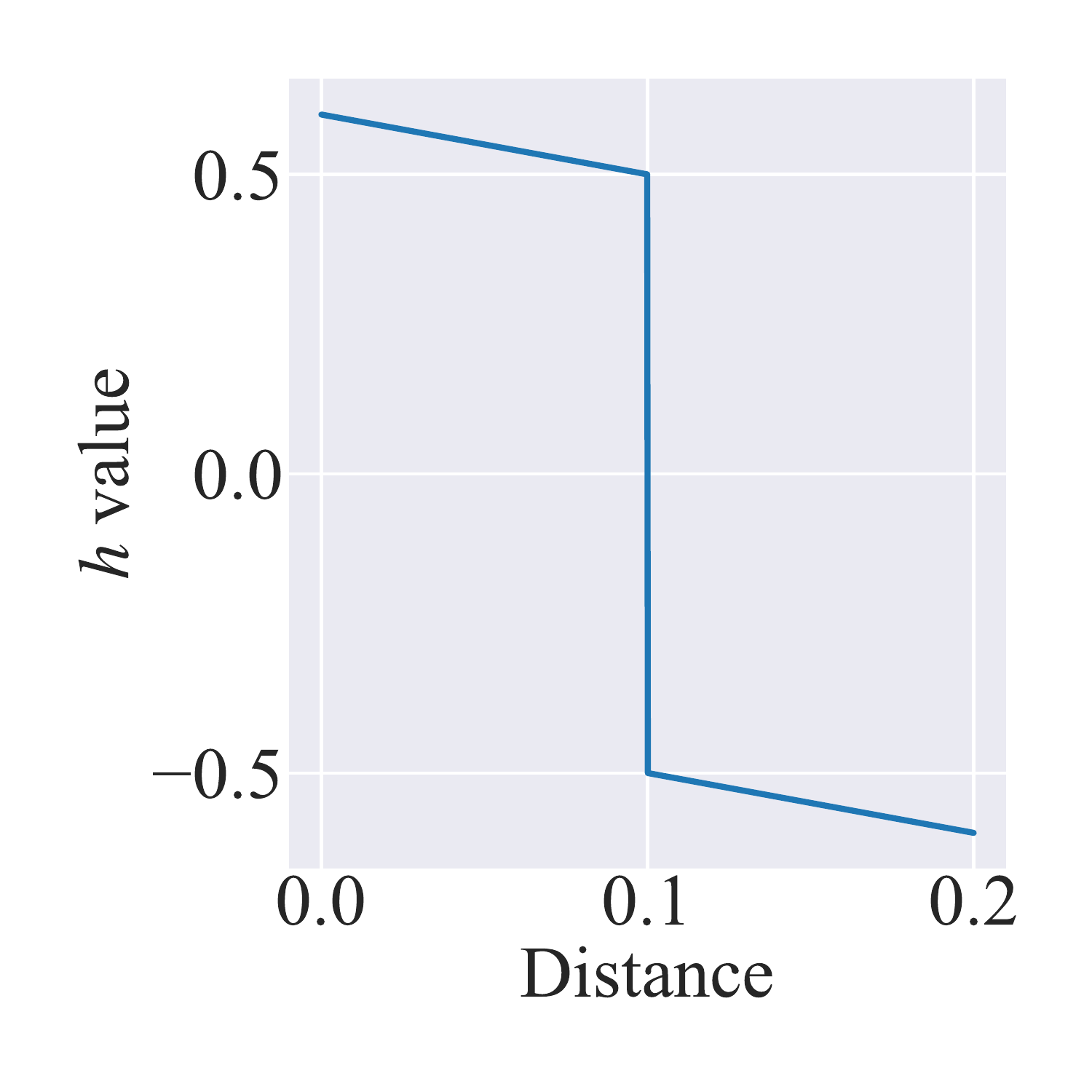}
    \caption{$h$ value with respect to distance.}
    \label{fig: h_d}
\end{figure}

The constraint function $h$ contains two parts for all environments except for \textsc{ConnnectSpread}, including agent-agent and agent-obstacles collisions. In the \textsc{ConnectSpread} environment, another constraint regarding the connectivity of the agent graph is considered. For the agent-agent collision, we use the $h$ function defined as
\begin{align}\label{eq: h_a}
    h_a(o_i) = 2r_a - \min_{j\in\mathcal N_i} \|p_i - p_j\| + \nu\mathrm{sign}\left(2r_a - \min_{j\in\mathcal N_i} \|p_i - p_j\|\right),
\end{align}
where $\mathrm{sign}$ is the sign function, and $\nu = 0.5$ in all our experiments. This represents a linear function w.r.t. the inter-agent distance with a discontinuity at the safe-unsafe boundary (\Cref{fig: h_d}). For the agent-obstacle collision, we use
\begin{align}\label{eq: h_o}
    h_o(o_i) = r_a + r_o - \min_{j\in\mathcal N_i^o} \|p_i - p_j\| + \nu\mathrm{sign}\left(r_a + r_o - \min_{j\in\mathcal N_i^o} \|p_i - p_j\|\right),
\end{align}
where $\mathcal N_i^o$ is the observed obstacle set of agent $i$. Then, the total $h$ function is defined as $h(o_i) = \max\{h_a(o_i), h_o(o_i)\}$ for environments except for \textsc{ConnectSpread}. For \textsc{ConnectSpread}, we also consider the connectivity constraint
\begin{align}
    h_c(o_i) = \max_i \min_{j\in\mathcal N_i^o} \|p_i - p_j\| - R' + \nu\mathrm{sign}\left(\max_i \min_{j\in\mathcal N_i^o} \|p_i - p_j\| - R' \right),
\end{align}
where $R' = 0.45$ is the required maximum distance for connected agents such that if the distance between two agents is larger than $R'$, they are considered disconnected. Note that this cost is only valid with agent number $N \leq 3$. For a larger number of agents, the second-largest eigenvalue of the graph Laplacian matrix can be used. Still, since we only use this environment with $3$ agents, we use this cost to decrease the complexity. Then, the total $h$ function of the \textsc{ConnectSpread} environment is defined as $h(o_i) = \max\{h_a(o_i), h_o(o_i), h_c(o_i)\}$. 

Two types of cost functions are used in the environments. The first type is the \textit{Target} cost used in the \textsc{Target} environment, which is defined as 
\begin{align}\label{eq: cost-target}
    l(x, u) = \frac{1}{N}\sum_{i=1}^N \left(0.01\|p_i - p_i^\mathrm{goal}\| + 0.001\mathrm{sign}\left(\mathrm{ReLU}(\|p_i - p_i^\mathrm{goal}\| - 0.01)\right) + 0.0001\|u_i\|^2\right).
\end{align}
The first term penalizes the agents if they cannot reach the goal, the second term penalizes the agents if they cannot reach the goal exactly, and the third term encourages small controls. The second type is the \emph{Spread} cost used in all other environments, defined as
\begin{align}\label{eq: cost-spread}
    l(x, u) = \frac{1}{N}\sum_{j=1}^N \min_{i\in\mathcal V_a}\left(0.01\|p_i - p_j^\mathrm{goal}\| + 0.001\mathrm{sign}\left(\mathrm{ReLU}(\|p_i - p_j^\mathrm{goal}\| - 0.01)\right)+ 0.0001\|u_j\|^2\right).
\end{align}
Instead of matching the agents to their preassigned goals, each goal finds its nearest agent and penalizes the whole team with the distance between them. In this way, the optimal policy of the agents is to cover all goals collaboratively. 

\subsubsection{Safe multi-agent MuJoCo environments}

We also test on the \textsc{Safe HalfCheetah(2x3)} and \textsc{Safe Coupled HalfCheetah(4x3)} tasks from the Safe Multi-Agent Mujoco benchmark suite \cite{gu2023safe}. Each agent controls a subset of joints and must cooperate to minimize the cost (which we take to be the negative of the reward in the original work) while avoiding violating safety constraints. The task is parametrized by the two numbers in the parentheses, where the first number denotes the number of agents, while the second number denotes the number of joints controlled by each agent. The goal for the \textsc{Safe HalfCheetah} and \textsc{Safe Coupled HalfCheetah} tasks is to maximize the forward velocity but avoid colliding with a wall in front that moves forward at a predefined velocity.

\textbf{Note: }
Although this is not a homogeneous MAS,
since each agent has the same control space (albeit with different dynamics), we can convert this into a homogeneous MAS by \textit{augmenting} the state space with a one-hot vector to identify each agent, then augmenting the dynamics to use the appropriate per-agent dynamics function.
This is the approach taken in the official implementation of Safe Multi-Agent Mujoco
from \citet{gu2023safe}. For more details, see \citet{gu2023safe}.

\subsection{Implementation details and hyperparameters}\label{app: implementation}

We parameterize the $z$-conditioned policy $\pi_\theta(o_i, z)$, cost-value function $V^l_\phi(x, z)$, and the constraint-value function $V^h_\psi(o_i, z)$ using graph transformers \citep{shi2020masked} with parameters $\theta$, $\phi$, and $\psi$, respectively. Note that the policy and the constraint-value function are decentralized and take only the local observation $o_i$ as input, while the cost-value function is centralized.
In each layer of the graph transformer, the node features are updated with $v_i' = W_1 v_i + \sum_{j\in\mathcal N_i}\alpha_{ij}(W_2 v_j + W_3 e_{ij})$, where $W_i$ are learnable weight matrices, and the $\alpha_{ij}$ is the attention weight between agent $i$ and agent $j$ computed as $\alpha_{ij} = \mathrm{softmax}(\frac{1}{\sqrt{c}}(W_4 x_i)^\top(W_5 x_j))$, where $c$ is the first dimension of $W_i$. In this way, the observation $o_i$ is encoded. If the environment allows $M$-hop information passing, we can apply the node feature update $M$ times so that agent $i$ can receive information from its $M$-hop neighbors.
After the information passing, the updated node features $v_i'$ are concatenated with the encoded $z$ vector $W_7 z$, then passed to another NN or a recurrent neural network (RNN) \citep{hausknecht2015deep} to obtain the outputs. $\pi_\theta$ and $V^h_\psi$ have the same structure as introduced above with different output dimensions because they are decentralized.
For the centralized $V^l(x, z)$, the averaged node features after information passing are concatenated with the encoded $z$ and passed to the final layer (NN or RNN) to obtain the global cost value for the whole MAS. 

When updating the neural networks, we follow the PPO \citep{schulman2017proximal} structure. First, we calculate the target cost-value function $V^l_\mathrm{target}$ and the target constraint-value function $V^h_\mathrm{target}$ using GAE estimation \citep{schulman2015high}, and then backpropagate the following mean-square error to update the value function parameters $\phi$ and $\psi$:
\begin{align}
    & \mathcal{L}_{V^l}(\phi) = \frac{1}{M}\sum_{k=1}^{M}\|V^l_\phi(x^k, z^k) - V^l_\mathrm{target}(x^k,z^k)\|^2, \\
    & \mathcal{L}_{V^h}(\psi) = \frac{1}{MN}\sum_{k=1}^{M}\sum_{i=1}^N\|V^h_\psi(o_i^k, z^k) - V^h_\mathrm{target}(o_i^k, z^k)\|^2,
\end{align}
where $M$ is the number of samples.
Then, we calculate the advantages $A_i$ for each agent with the total value function $V_i(x, z) = \max\{V^l_\phi(x, z) - z, V^h_\psi(o_i, z)\}$ following the same process as in PPO by replacing $V^l$ with $V$, and backpropagate the following PPO policy loss to update the policy parameters $\theta$:
\begin{equation}
    \begin{aligned}
        \mathcal L_\pi(\theta) = \frac{1}{MN}\sum_{k=1}^M\sum_{i=1}^N\Bigg[\min\Bigg\{\frac{\pi_\theta(o_i^k, z^k)}{\pi_\mathrm{old}(o_i^k, z^k)}A_i(x^k, z^k), 
        \mathrm{clip}\left(\frac{\pi_\theta(o_i^k, z^k)}{\pi_\mathrm{old}(o_i^k, z^k)}, 1-\epsilon_\mathrm{clip}, 1 + \epsilon_\mathrm{clip}\right)A_i(x^k, z^k)\Bigg\}\Bigg].
    \end{aligned}
\end{equation}

Most of the hyperparameters of \texttt{Def-MARL} are shared with \texttt{Penalty} and \texttt{Lagr}. The values of the share hyperparameters are provided in \Cref{tab: params-shared}. 

\begin{table}[h]
    \centering
    \caption{Shared hyperparameters of \texttt{Def-MARL}, \texttt{Penalty}, and \texttt{Lagr}.}
    \label{tab: params-shared}
    \begin{tabular}{lc|lc}
        \toprule
        Hyperparameter & Value & Hyperparameter & Value \\
        \midrule
        policy GNN layers & 2 & RNN type & GRU \\
        massage passing dimension & 32 & RNN data chunk length & 16\\
        GNN output dimension & 64 & RNN layers & 1 \\
        number of attention heads & 3 & number of sampling environments & 128 \\
        activation functions & ReLU & gradient clip norm & 2 \\
        GNN head layers & (32, 32) & entropy coefficient & 0.01 \\
        optimizer & Adam & GAE $\lambda$ & 0.95 \\
        discount $\gamma$ & 0.99 & clip $\epsilon$ & 0.25 \\
        policy learning rate & 3e-4 & PPO epoch & 1 \\
        $V^l$ learning rate & 1e-3 & batch size & 16384 \\
        network initialization & Orthogonal & layer normalization & True \\
        \bottomrule
    \end{tabular}
\end{table}

Apart from the shared hyperparameters, \texttt{Def-MARL} has additional hyperparameters, as shown in \Cref{tab: params-Def-MARL}. In addition, $z_\mathrm{min}$ and $z_\mathrm{max}$ are the lower and upper bounds of $z$ while sampling $z$ in training. Since $z_\mathrm{min}$ represents an estimate of the minimum cost incurred by the MAS, we set it to a small negative number $-0.5$. We set $z_\mathrm{max}$ differently depending on the complexity of the environment. For MPE, with maximum simulation timestep $T$, we estimate it in the MPE environments using the following equation:
\begin{align}
    z_\mathrm{max} &= \tilde{l}_\mathrm{max} * T, \\
    \tilde{l}_\mathrm{max} &= \mathrm{initdist}_{\max} \, w_\mathrm{distance} + w_\mathrm{reach} + u_\mathrm{max} w_\mathrm{control},
\end{align}
where $\tilde{l}_\mathrm{max}$ is a conservative estimate of the maximum cost $l$. This is conservative in the sense that this reflects the case where
1) the agents and goals are initialized with the maximum possible distance ($\mathrm{initdist}_{\max}$);
2) the agents do not reach their goal throughout their trajectory;
3) the agents incur the maximum control cost for all timesteps.
$w_\mathrm{distance}$, $w_\mathrm{reach}$, and $w_\mathrm{control}$ denote the corresponding weights of the different cost terms in the cost function $l$ in \eqref{eq: cost-target} and \eqref{eq: cost-spread}. For the multi-agent MuJoCo environments, we first train the agents with (unconstrained) MAPPO with different random seeds, record the largest cost incurred, double it, and then use that as $z_\mathrm{max}$.

\begin{table}[h]
    \centering
    \caption{Hyperparameters of \texttt{Def-MARL}.}
    \label{tab: params-Def-MARL}
    \begin{tabular}{l|c}
        \toprule
        Hyperparameter & Value \\
        \midrule
        $V^h$ GNN layers & 2 for \texttt{ConnectSpread}, 1 for others \\
        $z$ encoding dimension & 8 \\
        outer problem solver & Chandrupatla's method \citep{chandrupatla1997new}\\
\bottomrule
    \end{tabular}
\end{table}

All the hyperparameters remain the same in all environments or are pointed out in the tables except for the training steps. The training step is $10^5$ in the \textsc{Target} and the \textsc{Spread} environments, $1.5\times 10^5$ in the \textsc{Line} environment, and $2\times 10^5$ in other MPE. For the Safe Multi-agent MuJoCo environments, we set the training step to $7\times 10^3$.

\subsection{Training curves}\label{app: training-curve}

\begin{figure}[t]
    \centering
    \includegraphics[width=\columnwidth]{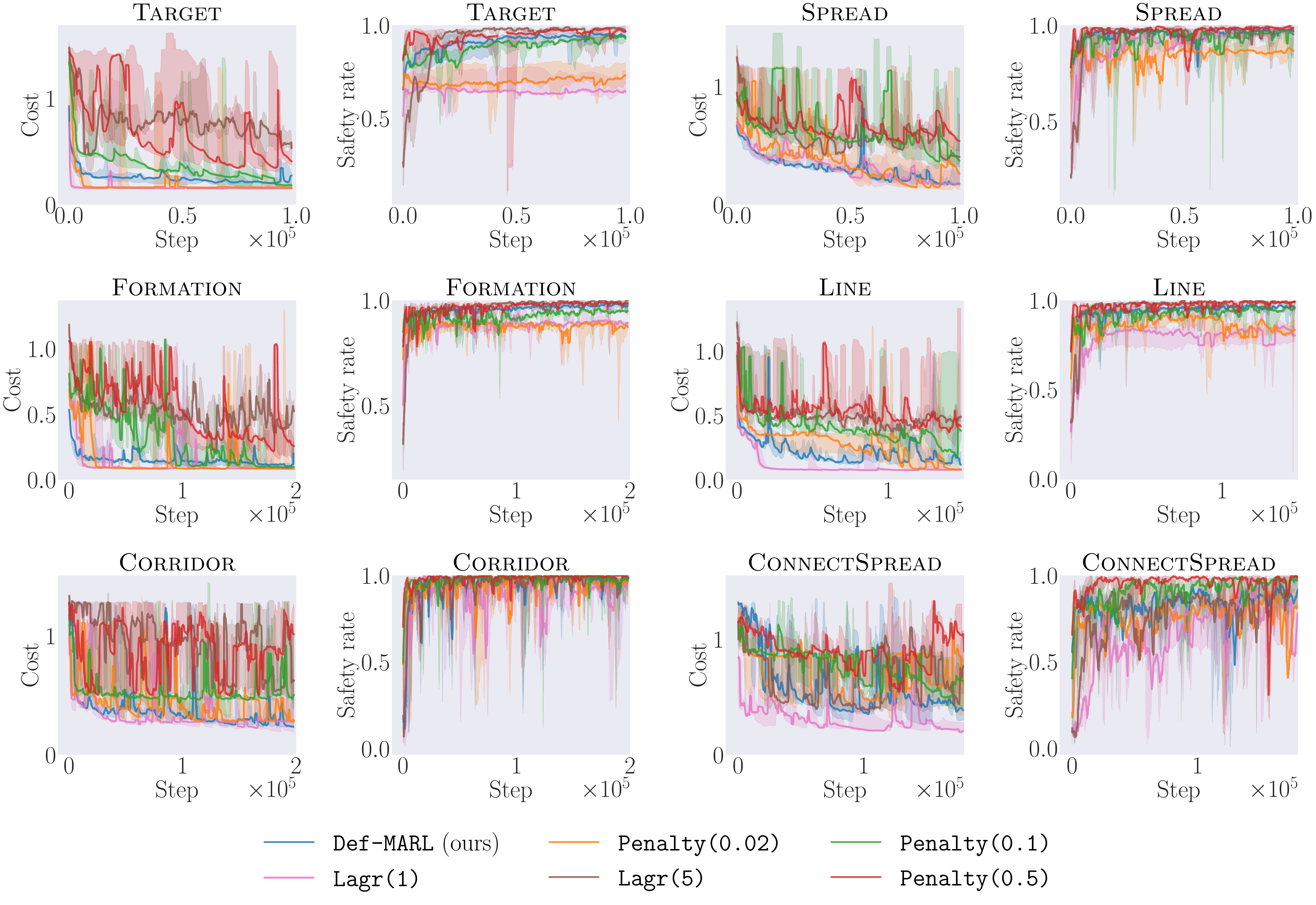}
    \caption{Cost and safety rate of \texttt{Def-MARL} and the baselines during training in MPE.}
    \label{fig: training_all}
\end{figure}

\begin{figure}[t]
    \centering
    \includegraphics[width=\columnwidth]{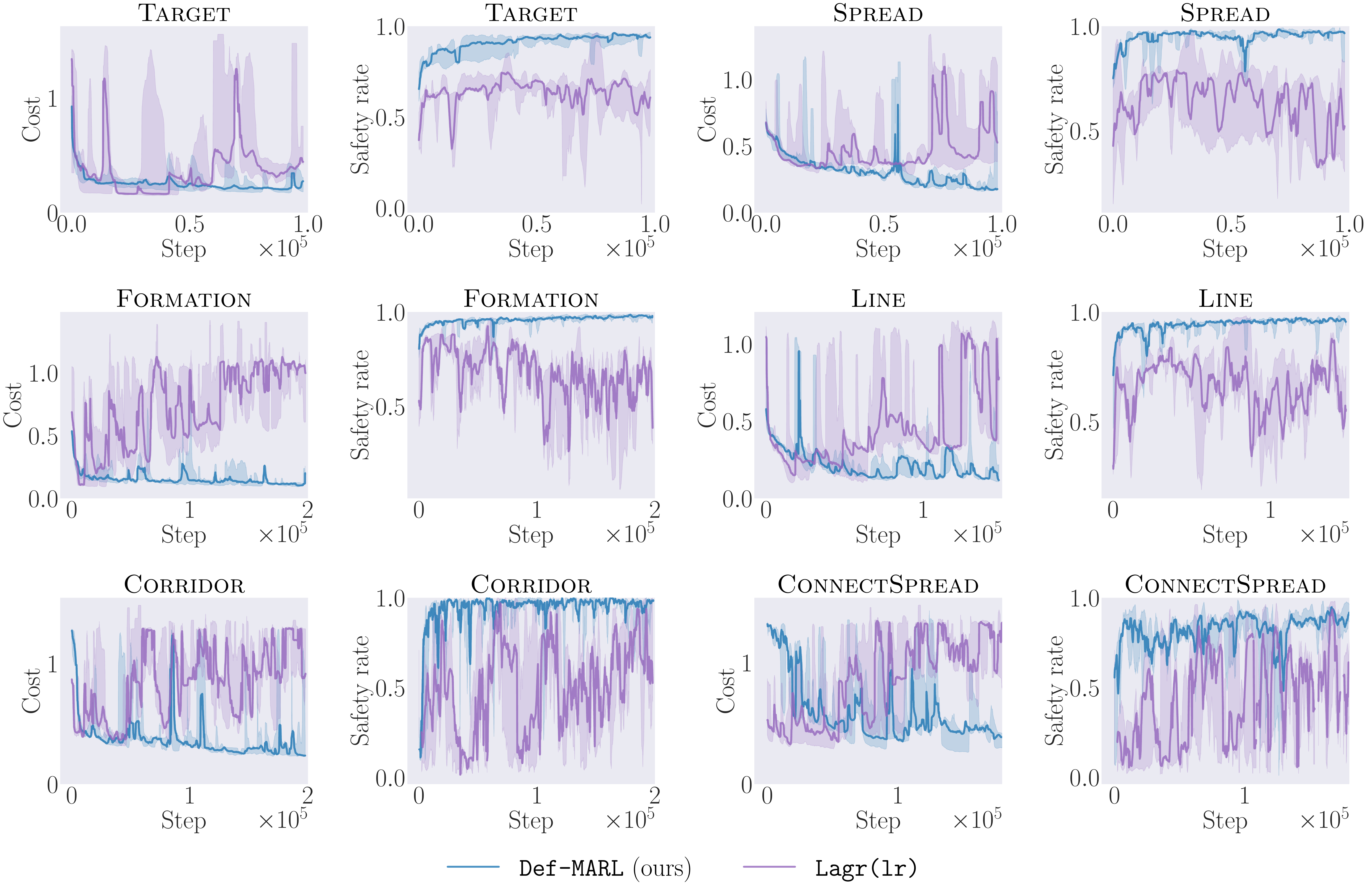}
    \caption{Cost and safety rate of \texttt{Def-MARL} and \texttt{Lagr(lr)} during training in MPE.}
    \label{fig: training_lagr_lr}
\end{figure}

\begin{figure}[t]
    \centering
    \includegraphics[width=\columnwidth]{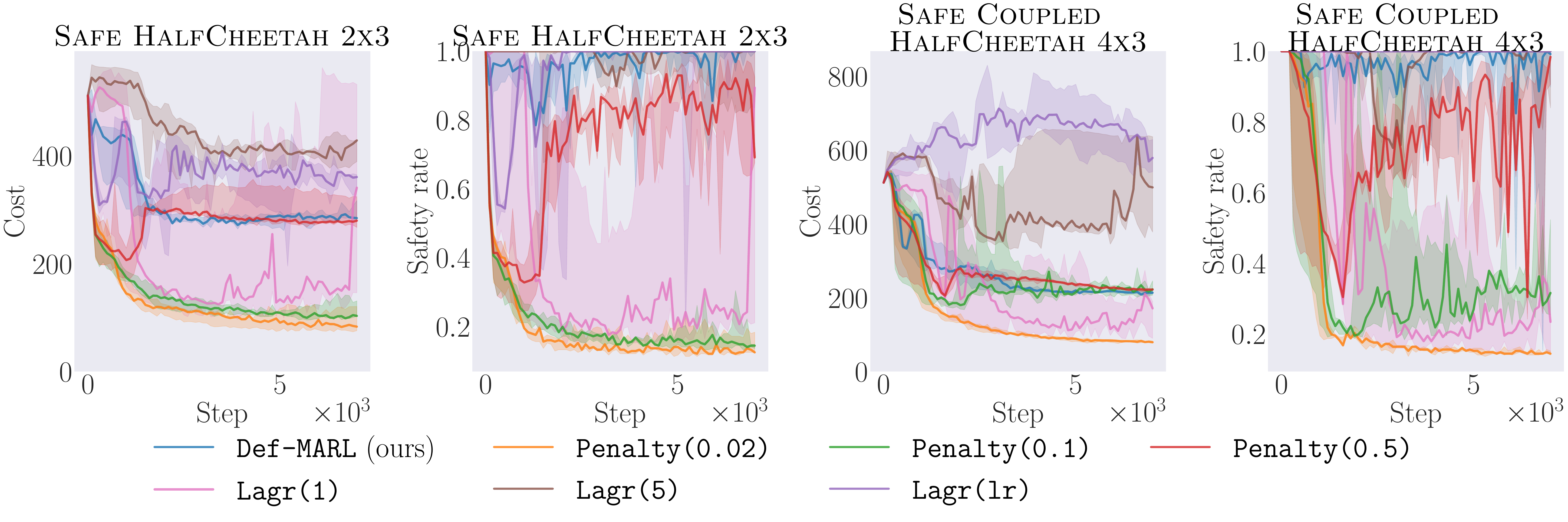}
    \caption{Cost and safety rate of \texttt{Def-MARL} and all baselines during training in Safe Multi-agent MuJoCo environments.}
    \label{fig: training_mujoco}
\end{figure}

To show the training stability of \texttt{Def-MARL}, we have shown the cost and safety rate of \texttt{Def-MARL} and \texttt{Lagr(lr)} during training in the \textsc{Target} and \textsc{Spread} environments in the main pages (\Cref{fig: training}). Due to page limits, we provide the plots for other environments here in \Cref{fig: training_all}, \Cref{fig: training_lagr_lr}, and \Cref{fig: training_mujoco}. The figures show that \texttt{Def-MARL} achieves stable training in all environments. Specifically, as shown in \Cref{fig: training_lagr_lr}, while \texttt{Lagr(lr)} suffers from training instability because the constraint violation threshold is zero (as discussed in \Cref{sec: EF}), \texttt{Def-MARL} is much more stable. 

\subsection{More comparison with the Lagrangian method}

\begin{figure}[t]
    \centering
    \includegraphics[width=\columnwidth]{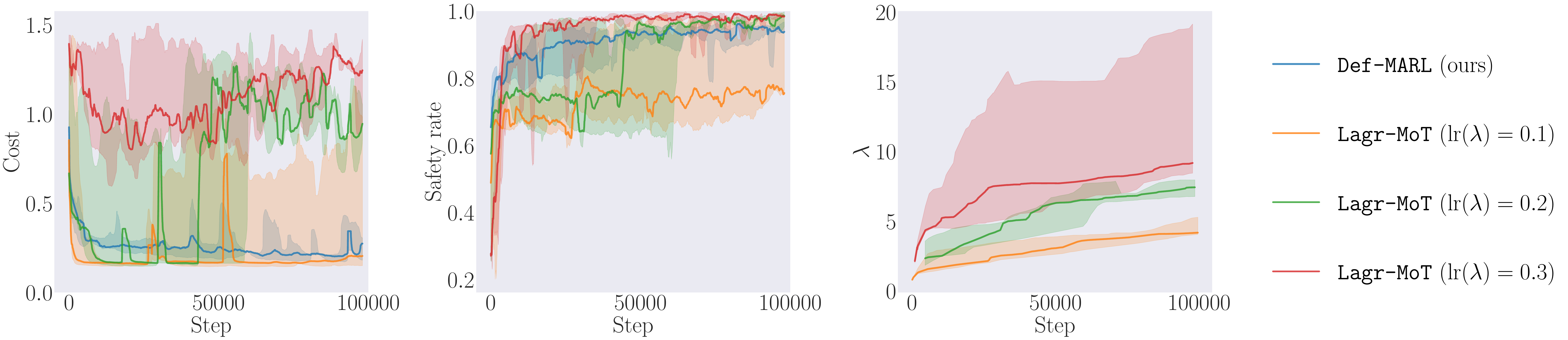}
    \caption{Cost and safety rate of \texttt{Def-MARL} and \texttt{Lagr-MoT} with different learning rates of $\lambda$ during training in the \textsc{Target} environment, and the $\lambda$ values during training.}
    \label{fig: training_lagr_max}
\end{figure}

In this section, we provide more comparisons between \texttt{Def-MARL} and the Lagrangian method, where we change the constraint-value function of the Lagrangian method from the sum-over-time (SoT) form to the max-over-time (MoT) form. Using the MoT form, the constraint-value function of the Lagrangian method becomes the same as the one used in \texttt{Def-MARL} (\Cref{eq: Vh-def}). We create $3$ more baselines using this approach with different learning rates (lr) of the Lagrangian multiplier $\lambda$, where $\mathrm{lr}(\lambda)\in\{0.1, 0.2, 0.3\}$. The baselines are called \texttt{Lagr-MoT}. We compare \texttt{Def-MARL} with the new baselines in the \textsc{Target} environment, and the results are presented in \Cref{fig: training_lagr_max}. We can observe that the Lagrangian method has very different performance with different learning rates of $\lambda$. With $\mathrm{lr}(\lambda) = 0.1$, the learned policy is unsafe, and with $\mathrm{lr}(\lambda) = 0.2$ or $0.3$, the training is unstable and the cost of the converged policy is much higher than \texttt{Def-MARL}. In addition, we also plot the $\lambda$ values during training in \Cref{fig: training_lagr_max}. It shows that $\lambda$ keeps increasing without converging to some value, which also suggests the instability of the Lagrangian method. 

\subsection{Sensitivity analysis on the choice of $z_\mathrm{max}$}

In \Cref{app: implementation}, we have introduced how to determine the sampling interval of $z$. Here, we perform experiments in the \textsc{Spread} environment to study the sensitivity of \texttt{Def-MARL} on the choice of $z_\mathrm{max}$. In this experiment, we scale the value of $z_\mathrm{max}$ used for sampling $z$, and denote by $z_\mathrm{max, orig}$ the original value used in the experiments in the main pages, i.e., $z_\mathrm{max} / z_\mathrm{max, orig} = 1.0$ uses the same value as in the main pages. We report the safety rates and the costs of the \texttt{Def-MARL} policies trained with different $z_\mathrm{max}$ in \Cref{tab: z-max-ablation}. 
We see that both safety and costs do not change much even when our estimate of the maximum cost $z_\mathrm{max}$ changes by up to $50\%$. 
If $z_\mathrm{max}$ is too large (e.g., $2 \, z_\mathrm{max, orig}$), the policy becomes too conservative because not enough samples of $z$ that are near $z^*$ are observed, reducing the sample efficiency.
On the other hand, when $z_\mathrm{max}$ is too small (e.g., $0.25 \, z_\mathrm{max, orig}$), there may be states where the optimal $z^*$ does not fall within the sampled range. This causes the rootfinding step to be inaccurate, as $V^h$ will be queried at values of $z$ that were not seen during training, resulting in safety violations.

\begin{table}[h]
    \centering
    \caption{Safety and Cost of \texttt{Def-MARL} policies trained with different $z_\mathrm{max}$.}
    \label{tab: z-max-ablation}
    \begin{tabular}{c|cc}
        \toprule
        $z_\mathrm{max} / z_\mathrm{max, orig}$ & Safety rate & Cost \\
        \midrule
        $0.25$ & $93.8\pm2.4$ & $0.152\pm0.100$\\
        $0.5$ & $98.0\pm1.4$ & $0.155\pm0.104$\\
        $1.0$ & $99.0\pm0.9$ & $0.162\pm0.144$\\
        $1.5$ & $99.0\pm0.0$ & $0.165\pm0.100$\\
        $2.0$ & $99.0\pm0.1$ & $0.228\pm0.109$\\
        \bottomrule
    \end{tabular}
\end{table}

\subsection{Effect of $z_i$ communication with more agents}

To test the effect of $z_i$ communication with more agents, we increase the number of agents to $512$ (at constant density) in \textsc{Target} environment during the test (\Cref{tab: communication-large-scale}).
Without communication, we only see a $0.3\%$ drop in safety rate going from $N=32$ to $N=512$. With $z$ communication, safety does not decrease with increased $N$.

\begin{table}[h]
    \centering
    \caption{Effect of $z_i$ communication in larger scale environments.}
    \label{tab: communication-large-scale}
    \begin{tabular}{c|cc|cc}
        \toprule
        \multirow{2}{*}{$N$} & \multicolumn{2}{c|}{No communication $(z \gets z_i)$} & \multicolumn{2}{c}{Communication ($z = \max_i z_i$)} \\
            & Safety rate & Cost & Safety rate & Cost \\
        \midrule
        32 & $99.8\pm0.2$ & $-0.387\pm0.029$ & $99.8\pm0.2$ & $-0.416\pm0.052$ \\
        128 & $99.6\pm0.4$ & $-0.408\pm0.015$ & $99.8\pm0.3$ & $-0.491\pm0.090$ \\
        512 & $99.5\pm0.3$ & $-0.410\pm0.009$ & $99.9\pm0.1$ & $-0.608\pm0.210$ \\
        \bottomrule
    \end{tabular}
\end{table}

\subsection{Code}

The code of our algorithm and the baselines are provided on our project website \url{https://mit-realm.github.io/def-marl/}.

\section{Hardware experiments}\label{app: hardware}

\subsection{Implementation details}

\textbf{Hardware platform. }
We perform hardware experiments using a swarm of Crazyflie (CF) 2.1 platform \cite{giernacki2017crazyflie}. We use two Crazyradios to communicate with the CFs and use the Lighthouse localization system to perform localization with four SteamVR Base Station 2.0. The computation is split into two parts: Onboard computation is performed on the CF microcontroller unit (MCU), and the offboard computation is performed on a laptop connected to the CFs over Crazyradio. We use the crazyswarm2 ROS2 package to communicate with the CFs, and a single ROS2 node for the off-board computations at 50Hz. 

\textbf{Control framework. }
Given a state estimation of the CFs using the Lighthouse system, the laptop inputs the states to the algorithm models (\texttt{Def-MARL}, \texttt{CMPC}, or \texttt{DMPC}), which then outputs the accelerations. Then, the laptop computes the desired next positions, velocities, and accelerations for the CFs, and sends the information to the CF MCU. After receiving the information, the CF MCU uses the onboard PID controller to track the desired position, velocity, and acceleration, which outputs the motor commands. 
\textit{Distributed} execution is simulated on the laptop by only giving each agent local information as input to their distributed policies.

\subsection{Tasks}

\textbf{\textsc{Corridor}. } The \textsc{Corridor} task uses the same cost and constraint function as the simulation \textsc{Corridor} environment, which has been introduced in \Cref{app: experiments-environments}.

\textbf{\textsc{Inspect}. } In this task, two CF drones need to work collaboratively to observe a moving target CF drone while maintaining collision-free with each other and the obstacles. The constraint function $h$ is defined the same as the simulation environments defined as $h(o_i) = \max\{h_a(o_i),h_o(o_i)\}$ with $h_a$ and $h_o$ defined in \eqref{eq: h_a} and \eqref{eq: h_o}. The cost function is given by
\begin{align}
    l(x, u) = l_\mathrm{visibility}(x) + 0.1\min_i l_\mathrm{dist}(x_i, x) + \frac{1}{N}\sum_{i=1}^N l_\mathrm{action}(u_i),
\end{align}
where $l_\mathrm{visibility}(x) = 0$ if the target is observable by at least one agent and $l_\mathrm{visibility}(x) = 0.01$ otherwise. For $l_\mathrm{dist}(x_i, x)$, we have $l_\mathrm{dist}(x_i, x) = 0$ if the target is observable by agent $i$, otherwise, it is defined as the distance between the farthest observed point on the agent-target line and the target. Finally, $l_\mathrm{action}(u_i)$ is defined as $l_\mathrm{action}(u_i) = 0.00001\|u_i\|^2$.

\subsection{Videos}

The videos of our hardware experiments are provided in the supplementary materials named `Corridor.mp4' and `Inspect.mp4'.

\section{Convergence}\label{app: convergence}

In this section, we analyze the convergence of the inner RL problem \eqref{eq: ef-macocp-inner} to a locally optimal policy.

Since we solve the inner RL problem \eqref{eq: ef-macocp-inner} in a centralized fashion, it can be seen as an instantiation of single-agent RL, but with a per-agent independent policy.
Define the augmented state $\tilde{x} \in \tilde{\mathcal{X}} \coloneqq \mathcal{X} \times \mathbb{R}$ as $[ x,\; z ]$, which follows the dynamics $\tilde{f} : \tilde{\mathcal{X}} \times \mathcal{U} \to \tilde{\mathcal{X}}$ defined as
\begin{equation}
    \tilde{f}\big([x^k, z^k], u^k\big) = \big[ f(x^k, u^k),\; z^k - l(x^k, u^k) \big].
\end{equation}
The inner RL problem \eqref{eq: ef-macocp-inner} can then be stated as
\begin{subequations}
\begin{align}
    \qquad\qquad \min_{\pi} \quad& \max_{k\geq0} h(x^k, \pi(\tilde{x}^k)) \\
    \subjto \quad
    & \tilde{x}^{k+1} = \tilde{f}(\tilde{x}^k, \tilde{\pi}( \tilde{x}^k)),  k\geq 0.
    \end{align}
\end{subequations}
This is an instance of a single-agent RL \textit{avoid} problem.
Consequently, applying the results from \cite[Theorem 5.5]{yu2022reachability} or \cite[Theorem 4]{so2024solving} gives us that the policy $\pi$ converges almost surely to a locally optimal policy.

\section{On the equivalence of the MASOCP and its epigraph form}

In \Cref{sec: EF}, we state that the Epigraph form \eqref{eq: ef} of a constrained optimization problem is equivalent to the original problem \eqref{eq:constr_problem}. This has been proved in \citet{so2023solving}. To make this paper more self-contained, we also include the proof here. 

\begin{proof}
    For a constrained optimization problem \eqref{eq:constr_problem}, its epigraph form \citep[pp 134]{boyd2004convex} is given by 
    \begin{subequations}
        \begin{align}
            \min_{\pi, z} \quad &z,\\
            \mathrm{s.t.} \quad & h(\pi)\leq 0,\label{eq: ef-constraint-1}\\
            &J(\pi)\leq z\label{eq: ef-constraint-2},
        \end{align}
    \end{subequations}
    where $z\in\mathbb R$ is an auxiliary variable. Here, \eqref{eq: ef-constraint-1} and \eqref{eq: ef-constraint-2} can be combined, which leads to the following problem:
    \begin{subequations}
        \begin{align}
            \min_{\pi, z} \quad &z,\\
            \mathrm{s.t.} \quad & \max\left\{h(\pi), J(\pi) - z\right\}\leq 0.
        \end{align}
    \end{subequations}
    Using this form, \citet[Theorem 3]{so2023solving} shows that the minimization of $x$ can be moved into the constraint, which yields
    \begin{subequations}
        \begin{align}
            \min_{z} \quad &z,\\
            \mathrm{s.t.} \quad & \min_\pi\max\left\{h(\pi), J(\pi) - z\right\}\leq 0.
        \end{align}
    \end{subequations}
    This is the same as \eqref{eq: ef}.
\end{proof}

\end{appendices}
\end{document}